\pdfoutput=1
\documentclass[11pt]{article}

\usepackage[preprint]{acl}

\usepackage{minted} 
\usepackage{float}
\usepackage{graphicx}
\usepackage{booktabs}
\usepackage{amsmath}
\usepackage{amssymb} 
\usepackage{caption}
\usepackage{times}
\usepackage{latexsym}
\usepackage{subfigure}                      
\usepackage[T1]{fontenc}
\usepackage[utf8]{inputenc}

\usepackage{microtype}

\usepackage{inconsolata}

\usepackage{siunitx}  
\usepackage{graphicx}
\usepackage{booktabs} 
\usepackage{multirow}
\usepackage{multicol}
\usepackage{enumitem}
\usepackage{algorithm}
\usepackage{algorithmic}
\usepackage{titlesec}
\usepackage{hyperref}

\titlespacing*{\section}
{0pt}                    % left margin
{0pt}                   % space before (vertical space above)
{0pt}                   % space after (vertical space below)

\titlespacing*{\subsection}{0pt}{0pt}{0pt}

\titlespacing*{\paragraph}{0pt}{0pt}{0pt}

\title{
PADBen: A Comprehensive Benchmark for Evaluating AI Text Detectors Against Paraphrase Attacks
}
\author{
\textbf{Yiwei Zha \textsuperscript{*,1}},
\textbf{Rui Min \textsuperscript{*,1}},
\textbf{Sushmita Shanu \textsuperscript{1}}
\\
\textsuperscript{*} Equal Contribution,
\textsuperscript{1} Khoury College of Computer Science, Northeastern University
}

\begin{document}
\maketitle
\begin{abstract}
While AI-generated text (AIGT) detectors achieve over 90\% accuracy on direct LLM outputs, they fail catastrophically against iteratively-paraphrased content. We investigate why iteratively-paraphrased text—itself AI-generated—evades detection systems designed for AIGT identification. Through intrinsic mechanism analysis, we reveal that iterative paraphrasing creates an intermediate laundering region characterized by semantic displacement with preserved generation patterns, which brings up two attack categories: paraphrasing human-authored text (authorship obfuscation) and paraphrasing LLM-generated text (plagiarism evasion). \\
To address these vulnerabilities, we introduce PADBen, the first benchmark systematically evaluating detector robustness against both paraphrase attack scenarios. PADBen comprises a five-type text taxonomy capturing the full trajectory from original content to deeply laundered text, and five progressive detection tasks across sentence-pair and single-sentence challenges. We evaluate 11 state-of-the-art detectors, revealing critical asymmetry: detectors successfully identify the plagiarism evasion problem but fail for the case of authorship obfuscation. Our findings demonstrate that current detection approaches cannot effectively handle the intermediate laundering region, necessitating fundamental advances in detection architectures beyond existing semantic and stylistic discrimination methods. For detailed code implementation, please see \href{https://github.com/JonathanZha47/PadBen-Paraphrase-Attack-Benchmark}{https://github.com/JonathanZha47/PadBen-Paraphrase-Attack-Benchmark}.
\end{abstract}
\section{Introduction}
Large Language Models (LLMs) like GPT-5, Claude-4, and Gemini-2.5 have achieved near-human quality in text generation \cite{openai2024gpt4technicalreport,geminiteam2025geminifamilyhighlycapable,kevian2024capabilitieslargelanguagemodels}. While enabling unprecedented automation across creative and academic domains, AI-generated text (AIGT) poses significant risks through malicious applications, including fabricating misinformation and automating spam \cite{leite2023detecting, Yeh2023EvaluatingIL}. This has spurred development of robust systems to differentiate human-authored from machine-generated text \cite{raid,bhattacharjee2023fightingfirechatgptdetect}.

A diverse ecosystem of AI text detectors has emerged, falling into two categories: zero-shot detectors like FastDetectGPT \cite{bao2024fastdetectgpt}, DetectGPT \cite{mitchell2023detectgptzeroshotmachinegeneratedtext}, GLTR \cite{gehrmann2019gltrstatisticaldetectionvisualization}, and Binoculars \cite{hans2024spottingllmsbinocularszeroshot}, which identify intrinsic statistical artifacts in synthetic text; and model-based detectors, including RADAR \cite{hu2023radarrobustaitextdetection} and OpenAI's RoBERTa classifier \cite{solaiman2019release}, fine-tuned on large datasets of human and AI content \cite{rezaei-etal-2024-clulab}. Recent research indicates that proprietary LLMs like GPT-4 and Qwen can be prompted to serve as effective detectors \cite{ji2025iknowbetterreally}.

Paraphrase attacks have emerged as the most effective evasion strategy. These attacks systematically reword AI-generated content while preserving semantic meaning, effectively ``laundering'' synthetic text to appear human-authored \cite{krishna2023paraphrasingevadesdetectorsaigenerated}. Advanced techniques like recursive paraphrasing significantly reduce detection performance while maintaining text quality \cite{sadasivan2023canaigeneratedtextreliably}. Unlike methods requiring deep technical expertise, paraphrasing is easily executed, causing state-of-the-art detectors' accuracy to plummet to near-random performance, creating severe risks from education to information security \cite{Weber_Wulff_2023, shportko-verbitsky-2025-paraphrasing}.

The prevalence of paraphrase attacks has exposed critical inadequacies in current evaluation frameworks for AIGT detection robustness. While existing benchmarks like RAID \cite{raid} provide comprehensive AIGT detection evaluation, they employ only single-step Dipper-based paraphrasing without systematic robustness assessment. Similarly, PARAPHRASUS \cite{michail2024paraphrasuscomprehensivebenchmark} evaluates paraphrase identification across multiple models using Classify, Min, and Max challenges on established NLP datasets. However, performing well on these challenges does not indicate robust adversarial defense, as these artificial scenarios focus on paraphrase detection rather than systematic evaluation of detector vulnerabilities to iterative evasion attacks. Neither framework addresses the critical gap: assessing detector performance against realistic, multi-iteration paraphrase-based attacks.

To address this gap, we introduce PADBen (\textbf{P}araphrase \textbf{A}ttack \textbf{D}etection \textbf{Ben}chmark), the first comprehensive benchmark to systematically evaluate AI text detectors against paraphrase attacks. Through dual representation space analysis, we observe that iterative paraphrasing creates an ``intermediate laundering region'' where texts undergo semantic drift while preserving generation patterns—a mechanism creating detection blind spots in current binary classification paradigms.

Based on this insight, we establish a five-type text taxonomy capturing the complete spectrum of authorship and paraphrasing dynamics: (1) \textbf{Type 1} - Human original text; (2) \textbf{Type 2} - LLM-generated text; (3) \textbf{Type 3} - Human-paraphrased original text; (4) \textbf{Type 4} - LLM-paraphrased original text; and (5) \textbf{Type 5} - Iteratively LLM-paraphrased LLM-generated text. Building upon this taxonomy, PADBen introduces five progressive detection tasks across two evaluation formats—single-sentence classification and sentence-pair recognition—designed to reflect realistic adversarial conditions.

Our key contributions are:
\begin{enumerate}[nosep]
    \item We are the first to systematically investigate paraphrase attack mechanisms through dual representation space analysis. We reveal that iterative paraphrasing creates an intermediate laundering region characterized by semantic displacement with preserved generation patterns, enabling two fundamentally distinct attack categories: authorship obfuscation (paraphrasing human-authored text) and plagiarism evasion (paraphrasing LLM-generated text);
    \item We propose a comprehensive five-type text taxonomy capturing both attack categories across their full trajectory from original content to deeply laundered text. We construct five progressive detection tasks evaluating detector robustness across sentence-pair and single-sentence formats, systematically assessing vulnerabilities to both authorship obfuscation and plagiarism evasion scenarios;
    \item We conduct extensive evaluations of 11 state-of-the-art detectors (4 zero-shot, 7 model-based), revealing critical asymmetry: paraphrase attacks do not universally defeat detection systems—outcomes depend on text origin.
\end{enumerate}
\section{Related Work}
\subsection{Paraphrase Attacks: A Primary Evasion Threat to AIGT Detection}
AIGT detectors face constant challenges from evasion techniques \cite{creo2025completeevasionzeromodification, lu2024largelanguagemodelsguided, zhou2024humanizingmachinegeneratedcontentevading, pudasaini-etal-2025-benchmarking}. Among various evasion strategies, paraphrase attacks—which employ language models to rewrite text while preserving semantic meaning—have emerged as a particularly potent threat \cite{Weber_Wulff_2023, sadasivan2023canaigeneratedtextreliably}. Research demonstrates that these attacks significantly compromise watermarking, zero-shot, and neural network-based detectors \cite{krishna2023paraphrasingevadesdetectorsaigenerated}. The study of paraphrase-based evasion is therefore essential for uncovering detector vulnerabilities and improving robustness, creating urgent need for rigorous evaluation frameworks.

\subsection{Existing Benchmarks and Gaps in Paraphrase Attack Evaluation}
Researchers have developed several major benchmarks targeting AIGT detection across diverse scenarios. RAID \cite{raid} encompasses over 6 million text generations from 11 language models across multiple domains, incorporating adversarial techniques including paraphrase attacks via Krishna et al.'s fine-tuned T5-11B models \cite{krishna2023paraphrasingevadesdetectorsaigenerated}. MAGE \cite{li2024magemachinegeneratedtextdetection} contributes 447k generations from 7 model families, emphasizing cross-domain and cross-model generalization. Complementary benchmarks address multilingual detection \cite{Macko_2023}, question-answering scenarios \cite{su2024hc3plussemanticinvarianthuman}, and scientific text discrimination \cite{mosca-etal-2023-distinguishing}.

Despite incorporating paraphrase attacks, these benchmarks treat paraphrasing as one perturbation among many rather than examining it as a distinct, evolving evasion pathway. This limited depth overlooks crucial challenges such as tracking degradation through iterative rewrites or assessing boundaries between laundering depths.

PARAPHRASUS \cite{michail2024paraphrasuscomprehensivebenchmark} targets paraphrase identification through three challenges across varying distributions: Classify (mixed), Minimize (0\%), and Maximize (100\% paraphrases). However, it focuses on paraphrase identification rather than adversarial robustness in AIGT detection. The extreme distributions may allow models to exploit dataset characteristics rather than generalizing to realistic scenarios.

Our work addresses these critical gaps by introducing PADBen, the first benchmark to systematically evaluate detector robustness against iterative paraphrase attacks in two distinct real-world scenarios: authorship obfuscation and plagiarism evasion. Unlike prior work treating paraphrasing as uniform single-step perturbations, PADBen evaluates progressive laundering across multiple iterations in both attack contexts. Through dual representation space analysis (Section~\ref{sec:mechanism}), we provide mechanistic insights into attack success patterns and identify critical vulnerabilities in current detection systems.
\section{How Do Paraphrase Attacks Intrinsically Work?}
\label{sec:mechanism}

\begin{figure*}[t]
    \centering
     \includegraphics[width=\textwidth]{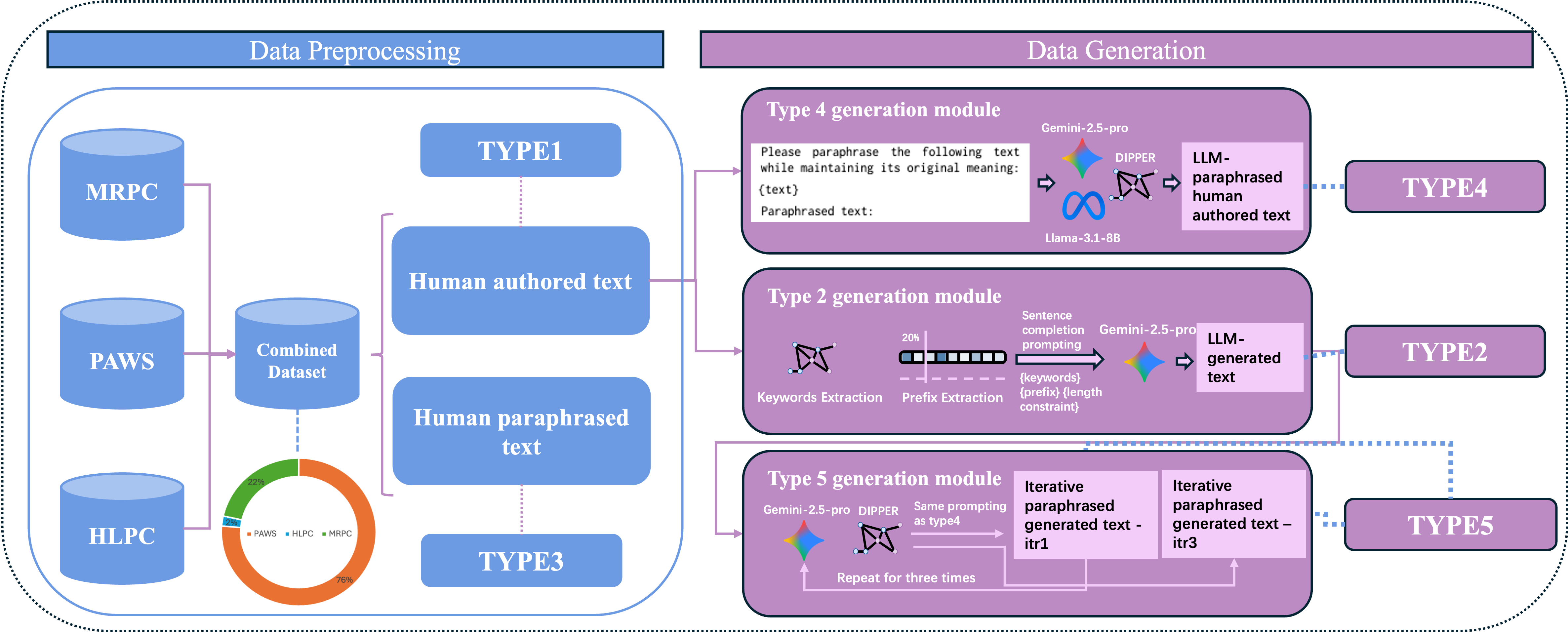}
    \caption{Overall pipeline for benchmark curation. Preprocessing details in Appendix~\ref{Appendix_A}, data generation in Appendix~\ref{appC:data_generation}.}
    \label{fig:overall_pipeline}
\end{figure*}

\textit{Since iteratively-paraphrased text is also AI-generated, why do paraphrase attacks evade AIGT detection systems?}

\noindent We hypothesize paraphrase attack effectiveness stems from unique representation space transformations. We formulate two testable hypotheses:

\textbf{Hypothesis 1:} Paraphrasing creates distinct semantic transformation differing from ``semantic equivalence'' prompting.

\textbf{Hypothesis 2:} Iterative paraphrasing increases coherence, deviating from LLM-generated patterns toward human-authored characteristics.

\noindent To test these hypotheses, we investigate how different prompting strategies (paraphrasing versus semantic equivalence) manifest in the model's representation space, and how iterative paraphrasing operations traverse this space over multiple iterations.
\subsection{Experimental Setup}

\textbf{Experiment 1:} We analyze three text categories in BGE-M3 embedding space: (1) human-authored, (2) LLM-generated via semantic equivalence prompts (GPT-4o), and (3) LLM-paraphrased human texts (GPT-4o). We apply PCA for 2D visualization while computing pairwise distances in full-dimensional space. K-means clustering ($k=3$) assesses separability (Appendix~\ref{app:experiment1}).

\noindent\textbf{Experiment 2:} We sample 100 texts each from human-authored and LLM-generated categories, performing 10 paraphrasing iterations using Qwen3-4B-Instruct. For each iteration, we extract: (1) paraphrased text, (2) final layer hidden states (4096-dim), and (3) BGE-M3 embeddings (1024-dim). We compute cosine, Euclidean, and Manhattan distances, applying PCA to centroid trajectories (Appendix~\ref{App:Experiment2}).

\subsection{Results and Analysis}
\label{finding_results}

\begin{table}[t]
\centering
\small
\caption{Pairwise semantic distances between text categories in BGE-M3 embedding space.}
\label{tab:pairwise_distances}
\begin{tabular}{@{}lccc@{}}
\toprule
\textbf{Comparison} & \textbf{Cosine} & \textbf{Eucl.} & \textbf{Manh.} \\
\midrule
Human $\leftrightarrow$ LLM-Gen. & 0.195 & 0.605 & 15.318 \\
Human $\leftrightarrow$ LLM-Para. & \textbf{0.068} & \textbf{0.355} & \textbf{8.991} \\
LLM-Gen. $\leftrightarrow$ LLM-Para. & 0.214 & 0.637 & 16.129 \\
\bottomrule
\end{tabular}
\end{table}

\textbf{Semantic Distinction Between Paraphrasing and Semantic Equivalence (Hypothesis 1)}

\noindent Table~\ref{tab:pairwise_distances} reveals LLM-paraphrased texts are \textbf{3.15$\times$ closer} to human originals (0.068 cosine similarity) than to LLM-generated texts (0.214), confirming paraphrased texts occupy an intermediate semantic region near human-authored content—\textbf{supporting Hypothesis 1}. On the other hand, Figure~\ref{fig:pca_clustering} shows an apparent paradox: While the above distance exhibits clear separability, 2D PCA results reveals substantial overlap. K-means clustering (Appendix~\ref{appB:experiment1_semantic_space}) produces mixed clusters across all text types. This is indicating semantic differences distribute across many dimensions rather than concentrating in low dimensionalities.

\begin{table}[t]
\centering
\small
\caption{Semantic distance (cosine and Euclidean) between iteratively paraphrased human text and two reference categories: original human-authored text and LLM-generated text in BGE-M3 embedding. Full table can be found in Table.\ref{tab:semantic_distances_detailed}}
\label{tab:semantic_distances}
\setlength{\tabcolsep}{4pt}
\begin{tabular}{@{}lcccccc@{}}
\toprule
\multirow{2}{*}{\textbf{Reference}} & \multirow{2}{*}{\textbf{Metric}} & \multicolumn{5}{c}{\textbf{Iteration}} \\
\cmidrule(lr){3-7}
& & \textbf{2} & \textbf{4} & \textbf{6} & \textbf{8} & \textbf{10} \\
\midrule
\multirow{2}{*}{\shortstack[l]{Human-\\Authored}} 
& Cosine & 0.085 & 0.107 & 0.122 & 0.128 & 0.134 \\
& Euclidean & 0.394 & 0.443 & 0.472 & 0.484 & 0.494 \\
\midrule
\multirow{2}{*}{\shortstack[l]{LLM-\\Generated}} 
& Cosine & 0.698 & 0.697 & 0.697 & 0.699 & 0.698 \\
& Euclidean & 1.180 & 1.180 & 1.179 & 1.181 & 1.180 \\
\bottomrule
\end{tabular}
\end{table}

\smallskip
\noindent \textbf{Semantic and Syntactic Impact of Iterative Paraphrasing (Hypothesis 2)}

\noindent Table~\ref{tab:semantic_distances} compares semantic distances: (1) human-authored text versus iteratively paraphrased human text shows progressive drift (cosine: 0.085→0.134), while (2) LLM-generated text versus iteratively paraphrased human text maintains stable distance (~0.698 across all iterations). These patterns \textbf{reject Hypothesis 2}—iterative paraphrasing increases distance from human texts while keeps a constant distance from LLM-generated text. 

To further examine the drift dynamics, Table~\ref{tab:iterative_distances} quantifies inter-iteration semantic changes across different representation spaces, revealing two key patterns: \textbf{(1) Progressive Semantic Shift:} Iterative-paraphrasing produce cumulative small semantic displacement for both human-authored and LLM-generated inputs. \textbf{(2) Representation-Dependent Drift Magnitude:} BGE-M3 embeddings exhibit larger inter-iteration displacement than Qwen3-4B hidden states.

These patterns reflect fundamental differences in what each representation captures—BGE-M3's contrastive training tracks \textit{semantic core} variations, while hidden states capture \textit{surface-level generation patterns} (lexical, syntactic, stylistic features). Thus, \textbf{iterative paraphrasing induces semantic shifts while preserving generation patterns}.

This mechanism enables two distinct attack scenarios: \textbf{(1) Authorship Obfuscation:} Human-authored text undergoing iterative paraphrasing maintains human-like stylistic markers despite semantic drift, creating detection blind spots that enable unauthorized appropriation of human writing. \textbf{(2) Plagiarism Detection Evasion:} LLM-generated text experiencing iterative paraphrasing preserves AI-like generation patterns while achieving sufficient semantic transformation to evade plagiarism detection systems, facilitating academic misconduct.

\begin{figure*}[t]
    \centering
     \includegraphics[width=\textwidth]{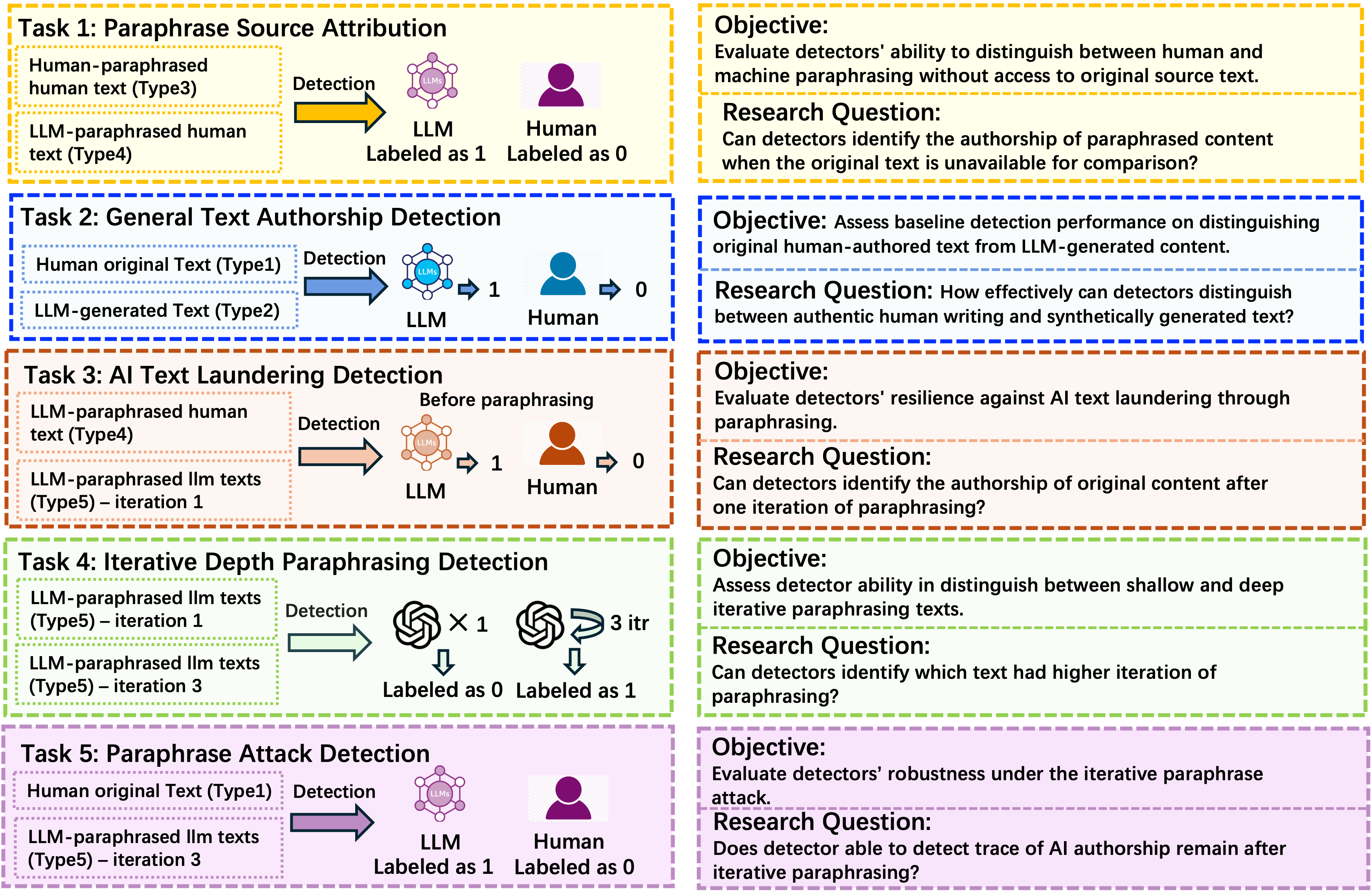}
    \caption{Overall Task introduction for the Benchmark. Task 1-5 measures the detector's different capabilities covering the robustness, performance when encountering the paraphrase attack. Detailed task specific can be found in Appendix.\ref{appC:task_intro}.}
    \label{fig:task_intro}
\end{figure*}
\noindent \textbf{Trajectory Analysis in Representation Space}

\noindent Figure~\ref{fig:trajectories_10itr} reveals both text origins converge toward similar regions with distinct patterns: hidden states show initial drift then oscillations; embeddings show gradual consistent drift. Directional convergence further support the existence of an "intermediate laundering region" in semantic space where texts deviate semantically from their origins while preserving generation characteristics. This region exhibits two properties: (1) \textit{universality}—accessible from both AI-generated and human-authored starting points; and (2) \textit{stability}—reliably reached via iterative paraphrasing.

\noindent \textbf{Summary}
\noindent Section~\ref{sec:mechanism} reveals a critical distinction in paraphrase attacks: iterative paraphrasing of human-authored text (authorship obfuscation) versus iterative paraphrasing of LLM-generated text (plagiarism evasion) represent fundamentally different risks, yet both exploit the same intermediate laundering region. Our mechanistic analysis demonstrates that regardless of origin, paraphrased texts converge toward this intermediate semantic space characterized by semantic displacement coupled with generation pattern preservation. This finding necessitates: (1) moving beyond binary human-versus-AIGT classification to capture how texts from different origins traverse through and occupy the intermediate region, and (2) incorporating multiple iterative paraphrasing depths to assess detector robustness as texts progressively enter this detection-resistant zone.

\begin{table*}[htbp]
\centering
\caption{Zero-shot Detectors Performance Summary}
\label{tab:appendix_zero_shot_complete}
\tiny
\setlength{\tabcolsep}{3pt}
\resizebox{\textwidth}{!}{%
\begin{tabular}{lll|cccc|cccc|cccc|cccc|cccc}
\toprule
\textbf{MODEL} & \textbf{CHALLENGE} & \textbf{METHOD}
& \multicolumn{4}{c|}{\textbf{Task 1}}
& \multicolumn{4}{c|}{\textbf{Task 2}}
& \multicolumn{4}{c|}{\textbf{Task 3}}
& \multicolumn{4}{c|}{\textbf{Task 4}}
& \multicolumn{4}{c}{\textbf{Task 5}} \\
 & & & AUC & T1 & T5 & T10 & AUC & T1 & T5 & T10 & AUC & T1 & T5 & T10 & AUC & T1 & T5 & T10 & AUC & T1 & T5 & T10 \\
\midrule
\multirow{5}{*}{BINOCULAR} 
 & sentence-pair   &      & 0.399 & 0.003 & 0.023 & 0.050 & 0.457 & 0.007 & 0.037 & 0.081 & 0.505 & 0.008 & 0.046 & 0.099 & 0.498 & \underline{\textcolor{blue}{0.011}} & 0.052 & 0.106 & 0.457 & 0.006 & 0.032 & 0.070 \\
 & single-sentence & exhaustive     & 0.439 & \underline{\textcolor{blue}{0.011}} & \underline{\textcolor{blue}{0.046}} & 0.080 & 0.579 & \underline{\textcolor{blue}{0.029}} & \underline{\textcolor{blue}{0.106}} & \underline{\textcolor{blue}{0.184}} & 0.500 & 0.008 & 0.050 & 0.101 & 0.486 & \underline{\textcolor{blue}{0.012}} & 0.050 & 0.098 & 0.428 & 0.009 & \underline{\textcolor{blue}{0.047}} & 0.087 \\
 & single-sentence & sampling\_30\% & 0.437 & \underline{\textcolor{blue}{0.010}} & \underline{\textcolor{blue}{0.046}} & 0.081 & 0.575 & \underline{\textcolor{blue}{0.031}} & \underline{\textcolor{blue}{0.109}} & \underline{\textcolor{blue}{0.188}} & 0.495 & 0.008 & 0.050 & 0.096 & 0.486 & \underline{\textcolor{blue}{0.010}} & \underline{\textcolor{blue}{0.051}} & 0.098 & 0.429 & 0.008 & \underline{\textcolor{blue}{0.050}} & 0.090 \\
 & single-sentence & sampling\_50\% & 0.443 & \underline{\textcolor{blue}{0.011}} & \underline{\textcolor{blue}{0.047}} & 0.084 & 0.583 & \underline{\textcolor{blue}{0.032}} & \underline{\textcolor{blue}{0.112}} & \underline{\textcolor{blue}{0.194}} & 0.499 & 0.008 & 0.051 & 0.102 & 0.487 & \underline{\textcolor{blue}{0.012}} & \underline{\textcolor{blue}{0.053}} & 0.099 & 0.434 & 0.009 & \underline{\textcolor{blue}{0.051}} & 0.092 \\
 & single-sentence & sampling\_80\% & 0.453 & \underline{\textcolor{blue}{0.010}} & \underline{\textcolor{blue}{0.049}} & 0.087 & 0.589 & \underline{\textcolor{blue}{0.034}} & \underline{\textcolor{blue}{0.116}} & \underline{\textcolor{blue}{0.195}} & 0.507 & 0.007 & 0.058 & 0.106 & 0.487 & \textbf{\textcolor{red}{0.016}} & \textbf{\textcolor{red}{0.060}} & \underline{\textcolor{blue}{0.104}} & 0.439 & \underline{\textcolor{blue}{0.012}} & \underline{\textcolor{blue}{0.053}} & 0.092 \\
\midrule
\multirow{5}{*}{\begin{tabular}{@{}l@{}}FAST\_DETECT\\\_GPT\end{tabular}} 
 & sentence-pair   &      & \underline{\textcolor{blue}{0.638}} & \textbf{\textcolor{red}{0.027}} & \textbf{\textcolor{red}{0.115}} & \underline{\textcolor{blue}{0.204}} & \underline{\textcolor{blue}{0.787}} & \underline{\textcolor{blue}{0.086}} & \underline{\textcolor{blue}{0.249}} & \underline{\textcolor{blue}{0.369}} & 0.503 & \underline{\textcolor{blue}{0.012}} & 0.055 & 0.108 & 0.476 & 0.005 & 0.036 & 0.081 & \underline{\textcolor{blue}{0.606}} & 0.023 & \underline{\textcolor{blue}{0.090}} & \underline{\textcolor{blue}{0.165}} \\
 & single-sentence & exhaustive     & \underline{\textcolor{blue}{0.573}} & 0.006 & 0.044 & \underline{\textcolor{blue}{0.099}} & \underline{\textcolor{blue}{0.665}} & 0.010 & 0.075 & 0.149 & 0.504 & 0.011 & 0.051 & 0.103 & 0.488 & 0.009 & 0.051 & 0.099 & \underline{\textcolor{blue}{0.568}} & 0.006 & 0.047 & \underline{\textcolor{blue}{0.103}} \\
 & single-sentence & sampling\_30\% & \underline{\textcolor{blue}{0.576}} & 0.006 & \underline{\textcolor{blue}{0.046}} & \underline{\textcolor{blue}{0.097}} & \underline{\textcolor{blue}{0.666}} & 0.009 & 0.078 & 0.154 & 0.502 & 0.011 & 0.049 & 0.096 & 0.490 & 0.007 & 0.046 & 0.091 & \underline{\textcolor{blue}{0.572}} & 0.005 & 0.045 & \underline{\textcolor{blue}{0.100}} \\
 & single-sentence & sampling\_50\% & \underline{\textcolor{blue}{0.581}} & 0.005 & 0.045 & \underline{\textcolor{blue}{0.098}} & \underline{\textcolor{blue}{0.672}} & 0.010 & 0.078 & 0.153 & 0.505 & 0.010 & 0.051 & 0.101 & 0.491 & 0.008 & 0.049 & 0.095 & \underline{\textcolor{blue}{0.577}} & 0.005 & 0.047 & \underline{\textcolor{blue}{0.102}} \\
 & single-sentence & sampling\_80\% & \underline{\textcolor{blue}{0.587}} & 0.004 & 0.040 & \underline{\textcolor{blue}{0.092}} & \underline{\textcolor{blue}{0.675}} & 0.008 & 0.076 & 0.151 & 0.514 & 0.007 & 0.045 & 0.100 & 0.488 & 0.011 & 0.049 & 0.095 & \underline{\textcolor{blue}{0.578}} & 0.005 & 0.048 & \underline{\textcolor{blue}{0.103}} \\
\midrule
\multirow{5}{*}{GLTR} 
 & sentence-pair   &      & 0.429 & \underline{\textcolor{blue}{0.007}} & 0.043 & 0.082 & 0.436 & 0.006 & 0.045 & 0.086 & \underline{\textcolor{blue}{0.529}} & 0.011 & \underline{\textcolor{blue}{0.060}} & \underline{\textcolor{blue}{0.116}} & \underline{\textcolor{blue}{0.514}} & \textbf{\textcolor{red}{0.016}} & \textbf{\textcolor{red}{0.057}} & \textbf{\textcolor{red}{0.113}} & 0.482 & \underline{\textcolor{blue}{0.012}} & 0.054 & 0.108 \\
 & single-sentence & exhaustive     & 0.459 & 0.006 & 0.032 & 0.068 & 0.480 & 0.004 & 0.021 & 0.056 & \underline{\textcolor{blue}{0.513}} & \underline{\textcolor{blue}{0.012}} & \underline{\textcolor{blue}{0.059}} & \underline{\textcolor{blue}{0.122}} & \underline{\textcolor{blue}{0.506}} & \textbf{\textcolor{red}{0.013}} & \textbf{\textcolor{red}{0.057}} & \textbf{\textcolor{red}{0.113}} & 0.488 & \underline{\textcolor{blue}{0.011}} & 0.047 & 0.091 \\
 & single-sentence & sampling\_30\% & 0.457 & 0.004 & 0.034 & 0.066 & 0.474 & 0.003 & 0.019 & 0.053 & \underline{\textcolor{blue}{0.519}} & \underline{\textcolor{blue}{0.012}} & \underline{\textcolor{blue}{0.065}} & \underline{\textcolor{blue}{0.124}} & \underline{\textcolor{blue}{0.502}} & \textbf{\textcolor{red}{0.014}} & \textbf{\textcolor{red}{0.056}} & \textbf{\textcolor{red}{0.109}} & 0.484 & \underline{\textcolor{blue}{0.012}} & 0.045 & 0.085 \\
 & single-sentence & sampling\_50\% & 0.461 & 0.005 & 0.036 & 0.068 & 0.480 & 0.004 & 0.022 & 0.057 & \underline{\textcolor{blue}{0.524}} & \underline{\textcolor{blue}{0.012}} & \underline{\textcolor{blue}{0.066}} & \underline{\textcolor{blue}{0.126}} & \textbf{\textcolor{red}{0.507}} & \textbf{\textcolor{red}{0.015}} & \textbf{\textcolor{red}{0.059}} & \textbf{\textcolor{red}{0.114}} & 0.489 & \underline{\textcolor{blue}{0.011}} & 0.049 & 0.089 \\
 & single-sentence & sampling\_80\% & 0.458 & 0.006 & 0.031 & 0.063 & 0.482 & 0.004 & 0.021 & 0.059 & \underline{\textcolor{blue}{0.523}} & \underline{\textcolor{blue}{0.011}} & \underline{\textcolor{blue}{0.062}} & \underline{\textcolor{blue}{0.117}} & \textbf{\textcolor{red}{0.509}} & \underline{\textcolor{blue}{0.013}} & \underline{\textcolor{blue}{0.059}} & \textbf{\textcolor{red}{0.111}} & 0.491 & 0.011 & 0.047 & 0.095 \\
\midrule
\multirow{5}{*}{RADAR} 
 & sentence-pair   &      & \textbf{\textcolor{red}{0.728}} & 0.004 & \underline{\textcolor{blue}{0.105}} & \textbf{\textcolor{red}{0.246}} & \textbf{\textcolor{red}{0.910}} & \textbf{\textcolor{red}{0.142}} & \textbf{\textcolor{red}{0.566}} & \textbf{\textcolor{red}{0.809}} & \textbf{\textcolor{red}{0.748}} & \textbf{\textcolor{red}{0.054}} & \textbf{\textcolor{red}{0.234}} & \textbf{\textcolor{red}{0.372}} & \textbf{\textcolor{red}{0.526}} & 0.010 & \underline{\textcolor{blue}{0.055}} & \underline{\textcolor{blue}{0.112}} & \textbf{\textcolor{red}{0.909}} & \textbf{\textcolor{red}{0.140}} & \textbf{\textcolor{red}{0.542}} & \textbf{\textcolor{red}{0.808}} \\
 & single-sentence & exhaustive     & \textbf{\textcolor{red}{0.648}} & \textbf{\textcolor{red}{0.038}} & \textbf{\textcolor{red}{0.190}} & \textbf{\textcolor{red}{0.345}} & \textbf{\textcolor{red}{0.793}} & \textbf{\textcolor{red}{0.063}} & \textbf{\textcolor{red}{0.313}} & \textbf{\textcolor{red}{0.567}} & \textbf{\textcolor{red}{0.633}} & \textbf{\textcolor{red}{0.016}} & \textbf{\textcolor{red}{0.080}} & \textbf{\textcolor{red}{0.160}} & \textbf{\textcolor{red}{0.511}} & 0.010 & \underline{\textcolor{blue}{0.052}} & \underline{\textcolor{blue}{0.104}} & \textbf{\textcolor{red}{0.797}} & \textbf{\textcolor{red}{0.062}} & \textbf{\textcolor{red}{0.310}} & \textbf{\textcolor{red}{0.562}} \\
 & single-sentence & sampling\_30\% & \textbf{\textcolor{red}{0.642}} & \textbf{\textcolor{red}{0.037}} & \textbf{\textcolor{red}{0.187}} & \textbf{\textcolor{red}{0.337}} & \textbf{\textcolor{red}{0.789}} & \textbf{\textcolor{red}{0.063}} & \textbf{\textcolor{red}{0.313}} & \textbf{\textcolor{red}{0.560}} & \textbf{\textcolor{red}{0.627}} & \textbf{\textcolor{red}{0.016}} & \textbf{\textcolor{red}{0.078}} & \textbf{\textcolor{red}{0.157}} & \textbf{\textcolor{red}{0.508}} & \underline{\textcolor{blue}{0.010}} & \underline{\textcolor{blue}{0.051}} & \underline{\textcolor{blue}{0.103}} & \textbf{\textcolor{red}{0.797}} & \textbf{\textcolor{red}{0.062}} & \textbf{\textcolor{red}{0.312}} & \textbf{\textcolor{red}{0.560}} \\
 & single-sentence & sampling\_50\% & \textbf{\textcolor{red}{0.644}} & \textbf{\textcolor{red}{0.036}} & \textbf{\textcolor{red}{0.181}} & \textbf{\textcolor{red}{0.337}} & \textbf{\textcolor{red}{0.789}} & \textbf{\textcolor{red}{0.060}} & \textbf{\textcolor{red}{0.302}} & \textbf{\textcolor{red}{0.559}} & \textbf{\textcolor{red}{0.628}} & \textbf{\textcolor{red}{0.016}} & \textbf{\textcolor{red}{0.078}} & \textbf{\textcolor{red}{0.155}} & \underline{\textcolor{blue}{0.506}} & 0.010 & 0.051 & \underline{\textcolor{blue}{0.102}} & \textbf{\textcolor{red}{0.795}} & \textbf{\textcolor{red}{0.060}} & \textbf{\textcolor{red}{0.300}} & \textbf{\textcolor{red}{0.556}} \\
 & single-sentence & sampling\_80\% & \textbf{\textcolor{red}{0.648}} & \textbf{\textcolor{red}{0.039}} & \textbf{\textcolor{red}{0.195}} & \textbf{\textcolor{red}{0.345}} & \textbf{\textcolor{red}{0.797}} & \textbf{\textcolor{red}{0.067}} & \textbf{\textcolor{red}{0.335}} & \textbf{\textcolor{red}{0.569}} & \textbf{\textcolor{red}{0.630}} & \textbf{\textcolor{red}{0.016}} & \textbf{\textcolor{red}{0.078}} & \textbf{\textcolor{red}{0.156}} & \underline{\textcolor{blue}{0.508}} & 0.010 & 0.051 & 0.102 & \textbf{\textcolor{red}{0.803}} & \textbf{\textcolor{red}{0.066}} & \textbf{\textcolor{red}{0.332}} & \textbf{\textcolor{red}{0.568}} \\
\bottomrule
\end{tabular}%
}
{\raggedright\small\textbf{Note:} AUC = AUC-ROC, T1 = TPR@1\%FPR, T5 = TPR@5\%FPR, T10 = TPR@10\%FPR. Best ({\color{red}\textbf{red bold}}) and second-best ({\color{blue}\underline{blue underlined}}) results are marked within each setup (sentence-pair, single-sentence exhaustive, sampling 30\%, 50\%, 80\%) for each task and metric.\par}
\end{table*}

\section{Methodology}

Section~\ref{sec:mechanism} reveals two different approaches in paraphrase attacks that existing benchmark did not address.
To systematically evaluate detection capabilities across both attack scenarios and the intermediate region they exploit, we develop a five-type text taxonomy. This taxonomy captures the full spectrum from original texts through the intermediate laundering region to deeply transformed content, enabling comprehensive evaluation of detector vulnerabilities against both paraphrase attack categories.
\subsection{Text Type Taxonomy}
We establish a five-category taxonomy:\\
\noindent\textbf{Type 1}: Human original text \\
\noindent\textbf{Type 2}: LLM-generated text \\
\noindent\textbf{Type 3}: Human-paraphrased human text \\
\noindent\textbf{Type 4}: LLM-paraphrased human text \\
\noindent\textbf{Type 5}: LLM-iteratively-paraphrased LLM text 

\noindent These clear categorized texts will help us build up our curation in task that mimiking most real world scenarios. Among them, the type 5 text will have both 1-iteration and 3-iteration. The detailed definition of them can be found in Appendix. \ref{appC:taxonomy_definition}.

\subsection{Data Preparation}
\paragraph{Source Data \& Data Preprocessing}~\\
Our benchmark leverages three established datasets: Microsoft Research Paraphrase Corpus (\textbf{MRPC})\cite{dolan2005automatically}, Human-LLM Paraphrase Corpus (\textbf{HLPC})\cite{lau2024hlpc}, and Paraphrase Adversaries from Word Scrambling (\textbf{PAWS})\cite{zhang2019paws}. We apply a cosine similarity filter (threshold: 0.85) to remove near-duplicates, and combined them to get 16233 human authored texts(Type1) and human-paraphrased human texts(Type3).

\paragraph{Generation Procedures}~\\
Figure \ref{fig:overall_pipeline} illustrates the detailed procedure of how raw data been preprocessed and how Type 2,4,5 texts are generated. As figure showed, we employed modularized generation pipeline for three categories in taxonomy:\\
\noindent\textbf{Type 2}: Sentence completion using Google Gemini-2.5-Pro\\
\noindent\textbf{Type 4}: Multi-model paraphrasing (DIPPER, Gemini-2.5-Pro, LLaMA-3-8B)\\
\noindent\textbf{Type 5}: Iterative paraphrasing with temperature scaling and convergence detection

\subsection{Quality Assurance}
To ensure the quality of our generated data, we examine the data quality by calculating three metrics: jaccard similarity, perplexity, and self-BLEU score. The jaccard similarity matrix across our text type taxonomy can be found in Figure.\ref{fig:jaccard_similarity}. Besides, Table.\ref{tab:quality_metrics} and \ref{tab:dataset_comparison} reveals that PADBen demonstrates superior dataset quality across three metrics. \\
Jaccard similarity confirms semantic preservation (0.798 for human paraphrases) while enabling controlled lexical divergence through iteration. \\
Perplexity analysis using GPT-2-XL and LLaMA-2-7B shows LLM-generated text exhibits lowest complexity (77.84/42.61), while human-authored and iteratively paraphrased texts achieve higher unpredictability (up to 109.32/50.23), indicating greater linguistic diversity.\\
Compared to RAID, PADBen achieves 62× higher intra-type diversity (self-BLEU: 0.222 vs 13.7) and 4.1-7.0× greater perplexity across evaluation models. This cross-model validation confirms PADBen generates more varied, complex content that effectively challenges detection systems.

\subsection{Detailed Task Introduction}

We design five progressive detection tasks to systematically assess AI text detection systems across varying complexity levels and attack scenarios. Each task targets specific vulnerabilities while leveraging our multi-type text dataset to evaluate detectors under increasingly sophisticated adversarial conditions. Figure. \ref{fig:task_intro} demonstrate the description of the five tasks. 

\noindent PADBen evaluates AI text detectors through five progressively challenging tasks:

\begin{enumerate}[nosep]
    \item \textbf{Task1: Paraphrase Source Attribution}:\\ Distinguish human-paraphrased (Type 3) from LLM-paraphrased (Type 4) text without original context
    
    \item \textbf{Task2: General Authorship Detection}:\\ Classify human original (Type 1) versus LLM-generated (Type 2) text—the baseline detection scenario
    
    \item \textbf{Task3: AI Text Laundering Detection}:\\ Identify whether paraphrased text originated from human (Type 4) or LLM sources (Type 5-1st) before transformation
    
    \item \textbf{Task4: Iterative Depth Detection}:\\ Distinguish shallow (Type 5-1st, 1 iteration) from deep paraphrasing (Type 5-3rd, 3 iterations) of the same LLM text
    
    \item \textbf{Task5: Paraphrase attack Detection}:\\ Classify human original (Type 1) versus maximally obfuscated AI text (Type 5-3rd)
\end{enumerate}

\noindent All tasks present sentence pairs in random order, with human/less-processed text labeled as 0 and LLM/more-processed text labeled as 1.
\begin{figure}[t]
    \centering
    \includegraphics[width=\columnwidth]{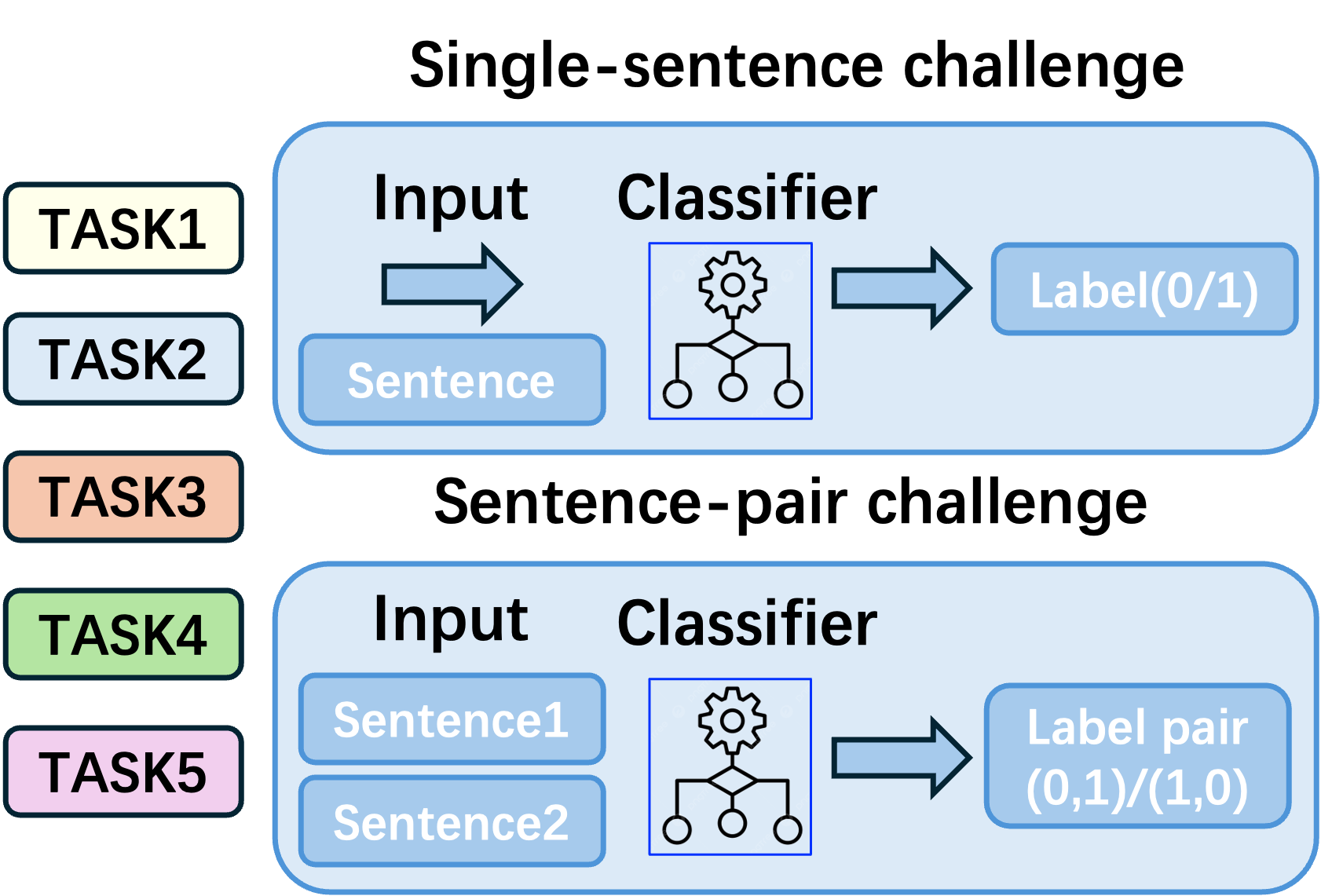}
    \caption{Two evaluation challenges: single-sentence classification and sentence-pair recognition. All five tasks are transformed into these two challenge formats. Detailed setup is provided in Appendix \ref{appD:evaluation_setup}.}
    \label{fig:evaluation_setup}
     \vspace{-1.5em} 
\end{figure}

\section{Evaluation Framework}

To comprehensively evaluate AI text detection capabilities, we examine two categories of detectors across multiple scenarios. Our selection covers both \textbf{Zero-shot Detectors} and \textbf{Model-based Detectors}, enabling a thorough assessment across methodological approaches.

\paragraph{Zero-shot detectors}~\\
Operate without task-specific training, relying on pre-existing linguistic/statistical patterns to distinguish human vs.~machine text. They include traditional statistical and rule-based systems while utilizing language models for extracting certain features. The state-of-the-art zero-shot detectors we evaluated include Binocular, Fast-Detect-GPT, GTLR, and RADAR.(Detailed technical setup can be found in Appendix.\ref{appE:zero-shot_setup}.

\paragraph{Model-based detectors}~\\
Leverage pre-trained language models or instruction-based LLM for classification, using internal representations learned from large-scale corpora to capture subtle differences between human and machine-generated content. We utilize few-shot and persona prompting strategies for both sentence-pair and single-sentence challenges. (Prompting details can see Appendix.\ref{appE:single-sentence_model_prompting}).

\subsection{Evaluation Setup}
We evaluate under two challenges: single-sentence classification and sentence-pair recognition with 5 different setups to reflect realistic use cases. Table.\ref{tab:setup_comparison} shows the main difference between 5 setups. Figure.\ref{fig:evaluation_setup} explains the inputs and outputs of the two challenges. Below are the listing for 5 setups:
\begin{enumerate}[nosep]
    \item \textbf{Single-Sentence Exhaustive:} Uses all available samples with balanced 50-50 distribution
    \item \textbf{Single-Sentence Sampling (30-70):} Random sampling with 30\% positive, 70\% negative
    \item \textbf{Single-Sentence Sampling (50-50):} Random sampling with balanced distribution
    \item \textbf{Single-Sentence Sampling (80-20):} Random sampling with 80\% positive, 20\% negative
    \item \textbf{Sentence-Pair Recognition:} Pairwise comparison tasks with random order presentation
\end{enumerate}
Details can be found from Appendix.\ref{appD:setup_1} to Appendix.\ref{appD:setup5}, illustrating the reason why we split into such settings and the algorithms for implementing. 

\begin{table*}[htbp]
\centering
\caption{Model-based Detectors Performance Summary}
\label{tab:appendix_model_based_complete}
\tiny
\setlength{\tabcolsep}{3pt}
\resizebox{\textwidth}{!}{%
\begin{tabular}{ll|cccc|cccc|cccc|cccc|cccc}
\toprule
\textbf{Model} & \textbf{Challenge}
& \multicolumn{4}{c|}{\textbf{Task 1}}
& \multicolumn{4}{c|}{\textbf{Task 2}}
& \multicolumn{4}{c|}{\textbf{Task 3}}
& \multicolumn{4}{c|}{\textbf{Task 4}}
& \multicolumn{4}{c}{\textbf{Task 5}} \\
 &  & \textbf{AUC} & \textbf{T1} & \textbf{T5} & \textbf{T10}
      & \textbf{AUC} & \textbf{T1} & \textbf{T5} & \textbf{T10}
      & \textbf{AUC} & \textbf{T1} & \textbf{T5} & \textbf{T10}
      & \textbf{AUC} & \textbf{T1} & \textbf{T5} & \textbf{T10}
      & \textbf{AUC} & \textbf{T1} & \textbf{T5} & \textbf{T10} \\
\midrule
\multirow{2}{*}{Claude-3.5-Haiku} 
& sentence-pair 
& \underline{\textcolor{blue}{0.623}} & \underline{\textcolor{blue}{0.016}} & \underline{\textcolor{blue}{0.078}} & \underline{\textcolor{blue}{0.155}} & 0.366 & 0.005 & 0.024 & 0.047 & 0.475 & \underline{\textcolor{blue}{0.010}} & 0.042 & 0.085 & 0.507 & \underline{\textcolor{blue}{0.010}} & 0.051 & 0.102 & 0.351 & 0.003 & 0.017 & 0.034 \\
& single-sentence
& \textbf{\textcolor{red}{0.514}} & \textbf{\textcolor{red}{0.012}} & \underline{\textcolor{blue}{0.058}} & \textbf{\textcolor{red}{0.115}} & \underline{\textcolor{blue}{0.535}} & \textbf{\textcolor{red}{0.018}} & \textbf{\textcolor{red}{0.089}} & \textbf{\textcolor{red}{0.162}} & 0.519 & \underline{\textcolor{blue}{0.022}} & \underline{\textcolor{blue}{0.069}} & 0.069 & 0.503 & \underline{\textcolor{blue}{0.011}} & 0.054 & 0.073 & \underline{\textcolor{blue}{0.545}} & \textbf{\textcolor{red}{0.049}} & \underline{\textcolor{blue}{0.112}} & \underline{\textcolor{blue}{0.112}} \\
\midrule
\multirow{2}{*}{DeepSeek-V2.5} 
& sentence-pair 
& 0.572 & 0.012 & 0.060 & 0.120 & 0.467 & 0.008 & 0.039 & 0.078 & \textbf{\textcolor{red}{0.519}} & \textbf{\textcolor{red}{0.011}} & \textbf{\textcolor{red}{0.057}} & \textbf{\textcolor{red}{0.113}} & \textbf{\textcolor{red}{0.514}} & \textbf{\textcolor{red}{0.011}} & \textbf{\textcolor{red}{0.054}} & \textbf{\textcolor{red}{0.108}} & 0.484 & 0.009 & 0.046 & 0.091 \\
& single-sentence
& 0.480 & 0.009 & 0.046 & 0.092 & 0.479 & 0.009 & 0.047 & 0.095 & \underline{\textcolor{blue}{0.531}} & 0.014 & \underline{\textcolor{blue}{0.069}} & \textbf{\textcolor{red}{0.138}} & 0.501 & 0.010 & 0.051 & 0.101 & 0.511 & 0.011 & 0.055 & 0.110 \\
\midrule
\multirow{2}{*}{Gemma-3-27B} 
& sentence-pair 
& 0.521 & 0.011 & 0.053 & 0.105 & \underline{\textcolor{blue}{0.506}} & \underline{\textcolor{blue}{0.010}} & 0.051 & \underline{\textcolor{blue}{0.102}} & 0.480 & \underline{\textcolor{blue}{0.010}} & 0.048 & 0.095 & 0.502 & \underline{\textcolor{blue}{0.010}} & 0.050 & 0.101 & 0.516 & \underline{\textcolor{blue}{0.010}} & 0.052 & 0.104 \\
& single-sentence
& 0.461 & 0.008 & 0.040 & 0.080 & 0.465 & 0.009 & 0.044 & 0.088 & 0.520 & 0.013 & 0.064 & 0.129 & 0.503 & 0.010 & 0.052 & 0.104 & 0.478 & 0.009 & 0.043 & 0.085 \\
\midrule
\multirow{2}{*}{Kimi-K2-Instruct} 
& sentence-pair 
& \textbf{\textcolor{red}{0.691}} & \textbf{\textcolor{red}{0.020}} & \textbf{\textcolor{red}{0.102}} & \textbf{\textcolor{red}{0.204}} & 0.431 & 0.007 & 0.036 & 0.071 & \underline{\textcolor{blue}{0.516}} & \textbf{\textcolor{red}{0.011}} & 0.053 & 0.106 & 0.487 & \underline{\textcolor{blue}{0.010}} & 0.048 & 0.096 & 0.441 & 0.006 & 0.030 & 0.060 \\
& single-sentence
& \underline{\textcolor{blue}{0.509}} & \textbf{\textcolor{red}{0.012}} & \textbf{\textcolor{red}{0.062}} & 0.096 & \textbf{\textcolor{red}{0.540}} & \underline{\textcolor{blue}{0.014}} & \underline{\textcolor{blue}{0.071}} & \underline{\textcolor{blue}{0.141}} & \textbf{\textcolor{red}{0.540}} & \textbf{\textcolor{red}{0.029}} & \textbf{\textcolor{red}{0.123}} & 0.123 & 0.503 & \underline{\textcolor{blue}{0.011}} & \textbf{\textcolor{red}{0.056}} & 0.063 & \textbf{\textcolor{red}{0.573}} & \underline{\textcolor{blue}{0.047}} & \textbf{\textcolor{red}{0.185}} & \textbf{\textcolor{red}{0.185}} \\
\midrule
\multirow{2}{*}{\begin{tabular}{@{}l@{}}Llama-4-Scout\\-17B\end{tabular}} 
& sentence-pair 
& 0.561 & 0.012 & 0.059 & 0.118 & 0.456 & 0.008 & 0.042 & 0.085 & \textbf{\textcolor{red}{0.519}} & \textbf{\textcolor{red}{0.011}} & \underline{\textcolor{blue}{0.054}} & \underline{\textcolor{blue}{0.109}} & 0.493 & \underline{\textcolor{blue}{0.010}} & 0.049 & 0.098 & 0.476 & 0.009 & 0.046 & 0.091 \\
& single-sentence
& 0.486 & 0.009 & 0.047 & 0.095 & 0.472 & 0.009 & 0.046 & 0.091 & 0.523 & 0.013 & 0.065 & \underline{\textcolor{blue}{0.131}} & \textbf{\textcolor{red}{0.509}} & \underline{\textcolor{blue}{0.011}} & \underline{\textcolor{blue}{0.055}} & \textbf{\textcolor{red}{0.111}} & 0.506 & 0.010 & 0.052 & 0.103 \\
\midrule
\multirow{2}{*}{Mistral-Nemo} 
& sentence-pair 
& 0.510 & 0.014 & 0.069 & 0.073 & 0.504 & \textbf{\textcolor{red}{0.011}} & \textbf{\textcolor{red}{0.057}} & 0.066 & 0.501 & \underline{\textcolor{blue}{0.010}} & 0.051 & 0.101 & 0.502 & \underline{\textcolor{blue}{0.010}} & 0.051 & 0.101 & \underline{\textcolor{blue}{0.518}} & \textbf{\textcolor{red}{0.012}} & \underline{\textcolor{blue}{0.059}} & \underline{\textcolor{blue}{0.118}} \\
& single-sentence
& 0.489 & 0.009 & 0.044 & 0.089 & 0.470 & 0.008 & 0.040 & 0.081 & 0.494 & 0.008 & 0.038 & 0.040 & 0.493 & 0.009 & 0.046 & 0.091 & 0.498 & 0.009 & 0.046 & 0.049 \\
\midrule
\multirow{2}{*}{\begin{tabular}{@{}l@{}}WizardLM-2\\-8x22B\end{tabular}} 
& sentence-pair 
& 0.558 & 0.012 & 0.059 & 0.119 & \textbf{\textcolor{red}{0.520}} & \textbf{\textcolor{red}{0.011}} & \underline{\textcolor{blue}{0.054}} & \textbf{\textcolor{red}{0.108}} & 0.508 & \underline{\textcolor{blue}{0.010}} & \underline{\textcolor{blue}{0.052}} & 0.103 & \underline{\textcolor{blue}{0.510}} & \underline{\textcolor{blue}{0.010}} & \underline{\textcolor{blue}{0.052}} & \underline{\textcolor{blue}{0.104}} & \textbf{\textcolor{red}{0.551}} & \textbf{\textcolor{red}{0.012}} & \textbf{\textcolor{red}{0.061}} & \textbf{\textcolor{red}{0.122}} \\
& single-sentence
& 0.501 & \underline{\textcolor{blue}{0.010}} & 0.050 & \underline{\textcolor{blue}{0.101}} & 0.505 & 0.011 & 0.053 & 0.105 & 0.505 & 0.013 & 0.045 & 0.045 & \underline{\textcolor{blue}{0.504}} & \textbf{\textcolor{red}{0.012}} & 0.049 & 0.049 & 0.499 & 0.009 & 0.047 & 0.048 \\
\bottomrule
\end{tabular}%
}
{\raggedright\small\textbf{Note:} AUC = AUC-ROC, T1 = TPR@1\%FPR, T5 = TPR@5\%FPR, T10 = TPR@10\%FPR. Single-sentence evaluation uses 50--50 sampling. Best ({\color{red}\textbf{red bold}}) and second-best ({\color{blue}\underline{blue underlined}}) results are marked within each setup (sentence-pair or single-sentence) for each task and metric.\par}
\end{table*}

\section{Results: Task-by-Task Performance Analysis}
We evaluate 4 zero-shot and 7 model-based detectors across five tasks. Evaluation results reveal systematic vulnerabilities aligned with our mechanistic understanding: both paraphrase attack categories—authorship obfuscation (paraphrasing human text) and plagiarism evasion (paraphrasing LLM text)—exploit the intermediate laundering region identified in Section~\ref{sec:mechanism}. We analyze performance patterns task-by-task, integrating findings from both detector categories. Tables~\ref{tab:appendix_zero_shot_complete} and~\ref{tab:appendix_model_based_complete} show complete results.

%\paragraph{Task-by-Task Performance Analysis}
\mbox{}\\
\textbf{Task 1 (Paraphrase Source Attribution):} Both detector categories struggle with absolute classification (AUC 0.46-0.52 single-sentence) but show improved sentence-pair performance. RADAR achieves best results (AUC 0.728 sentence-pair, 0.648 exhaustive), followed by Kimi-K2-Instruct (AUC 0.691 sentence-pair). This difficulty directly validates our finding that both human and LLM paraphrasing converge toward the intermediate laundering region (Section~\ref{finding_results}), making source attribution challenging while preserving comparative signals.

\noindent \textbf{Task 2 (General Authorship Detection):} RADAR dominates with AUC 0.910 (sentence-pair) and 0.797 (exhaustive), exploiting the clear semantic separation between human and LLM text (0.195 cosine similarity, Table~\ref{tab:pairwise_distances}). Model-based detectors underperform substantially (best: Kimi AUC 0.540), suggesting instruction-following models cannot exploit representation space differences without fine-tuning. Most other detectors show near-random performance (AUC < 0.6).

\noindent \textbf{Task 3 (AI Text Laundering Detection):} Performance collapses across all detectors, empirically validating the intermediate laundering region's detection blind spot. RADAR maintains moderate sentence-pair capability (AUC 0.748) but single-sentence performance degrades to near-random (0.50-0.63). Model-based detectors show inverted patterns with Kimi achieving best single-sentence results (AUC 0.540), suggesting sensitivity to absolute laundering signatures. However, all performance remains barely above chance, confirming the authorship obfuscation attack successfully masks source attribution once texts enter the intermediate region.

\noindent \textbf{Task 4 (Iterative Depth Detection):} Universal failure across all detectors (AUC 0.487-0.529) validates our trajectory analysis showing oscillatory movement within the intermediate region. Neither zero-shot nor model-based approaches can extract depth information, as intermediate laundering region eliminates iteration-specific signatures while maintaining stable generation patterns.

\noindent \textbf{Task 5 (Paraphrase Attack Detection):} RADAR demonstrates strong performance (AUC 0.909 sentence-pair, 0.803 exhaustive), confirming our finding that deeply laundered AI text maintains stable distance from human originals despite semantic drift (Table~\ref{tab:semantic_distances}). This validates the plagiarism evasion attack mechanism: iteratively paraphrased LLM text preserves AI-like generation patterns detectable against human baselines. Model-based detectors show modest capability (Kimi AUC 0.573 single-sentence) but cannot match zero-shot performance.
\vspace{-3pt}
\paragraph{Common Observations} 
\mbox{}\\
\textbf{Two Attack Categories Validated:} Task 3's failure (AUC 0.748) versus Task 5's success (AUC 0.909) empirically confirms the distinction between authorship obfuscation and plagiarism evasion attacks. Both exploit the intermediate laundering region but produce different detection signatures—source attribution becomes impossible (Task 3) while human-vs-laundered-AI discrimination remains feasible (Task 5).\\
\textbf{Intermediate Laundering Region Properties:} Task 3's catastrophic collapse ($\text{AUC } 0.9{+} \rightarrow 0.6 \text{--} 0.7$) and Task 4's universal failure ($\text{AUC} \approx 0.5$) validate this region's universality (accessible from both origins) and stability (reliably reached via iteration), creating fundamental blind spots for source attribution and depth discrimination.\\
\textbf{Semantic Drift with Pattern Preservation:} Divergent Task 3/5 performance confirms iterative paraphrasing shifts semantic positioning while preserving generation patterns—artifacts survive multiple iterations enabling Task 5 detection, yet become uninformative for Task 3 source attribution.\\
\textbf{Representation Space Asymmetry:} RADAR's superiority reflects semantic variations distributing across high-dimensional embedding space while generation patterns concentrate in low-dimensional features—zero-shot methods leverage the former, instruction-following models struggle with the latter.\\
\textbf{Evaluation Robustness:} Single-sentence sampling methods show stability ($\text{variation} \leq 0.02~\text{AUC}$), confirming stylistic discrimination over content memorization. Zero-shot detectors benefit from sentence-pair evaluation ($(+0.1\text{--}0.2~\text{AUC})$), while model-based detectors show inconsistent patterns.

\section{Conclusion}
This work reveals a fundamental challenge in AI text detection: paraphrase attacks do not universally defeat detection systems—outcomes critically depend on text origin.Through dual representation space analysis, we identify the intermediate laundering region as the key mechanism enabling two distinct attack categories: authorship obfuscation and plagiarism evasion. These attacks exploit this region differently—iteratively paraphrased LLM text preserves detectable generation artifacts, while iteratively paraphrased human text maintain the human tone that confound source attribution.
To systematically evaluate these vulnerabilities, we introduce PADBen, the first benchmark assessing detector robustness against both attack scenarios. PADBen provides the research community with: (1) a comprehensive five-type text taxonomy capturing the full attack trajectory, (2) five progressive detection tasks across realistic conditions, and (3) mechanistic insights into why current binary classifiers fail within the intermediate region. Our evaluation of 11 state-of-the-art detectors confirms this asymmetry: plagiarism evasion remains detectable (RADAR AUC 0.909), while authorship obfuscation collapses detection to near-random performance (AUC 0.526-0.748).

\section{Ethics Statement}
We use only publicly available datasets and pre-trained models in this study, all of which are accessed and utilized strictly for research purposes. The use of these resources complies with their original licenses and terms of access. No personally identifiable or sensitive information is present in any of the data used.

Our code will be released under the MIT license to support transparency and reproducibility.

\section{Limitations}

While PADBen provides comprehensive evaluation of paraphrase attack robustness, several aspects could be enhanced in future iterations:

\textbf{Experimental Controls.} Our Experiment 2 trajectory analysis could benefit from stricter variable control, particularly maintaining consistent text length across paraphrasing iterations and expanding sample sizes beyond 100 texts per category. These enhancements would strengthen statistical power and eliminate potential length-based confounds. However, controlling paraphrased text length while preserving semantic content presents inherent trade-offs, as natural paraphrasing often alters length. We prioritized semantic fidelity over length consistency to reflect realistic paraphrase attack scenarios.

\textbf{Taxonomy Coverage.} Our five-type taxonomy focuses on core paraphrase attack scenarios but could expand to include additional variants. Specifically, extending Type 5 beyond 3 iterations (e.g., 5-10 iterations) and introducing intermediate iteratively-paraphrased human texts variants (Type 4 with 2-10 iterations) would enable finer-grained robustness assessment. Due to computational constraints—generating and validating 16,233 texts across multiple iteration depths requires substantial computing costs and processing time—we prioritized depth ranges that capture critical transition points into the intermediate laundering region while maintaining dataset quality.

\textbf{Detector Optimization.} We evaluate zero-shot detectors using default configurations without fine-tuning on PADBen data. While this approach assesses out-of-the-box robustness, adapted implementations could potentially improve performance. Fine-tuning experiments would require extensive hyperparameter search across multiple detectors and task configurations, which was beyond our resource constraints. Nevertheless, our results establish baseline performance against which adapted methods can be compared.

\textbf{Evaluation Comprehensiveness.} Current sentence-pair tasks exclusively include paraphrased text in at least one position. Incorporating additional pairs without paraphrasing (e.g., human-original vs. LLM-generated-original) would provide more diverse evaluation scenarios and test whether detectors rely on paraphrasing artifacts versus genuine authorship signals. Future work should integrate such controls to eliminate potential evaluation biases, though our primary focus remains paraphrase attack robustness rather than general authorship detection.

These limitations represent opportunities for enhancement rather than fundamental flaws. PADBen's current design prioritizes realistic attack scenarios, mechanistic insights, and comprehensive detector evaluation within practical resource constraints, providing a solid foundation for future extensions addressing these aspects.

\bibliography{custom}
\newpage
\appendix
\section{Overall Data Processing}
\label{Appendix_A}
\begin{figure*}[t]
    \centering
     \includegraphics[width=\textwidth]{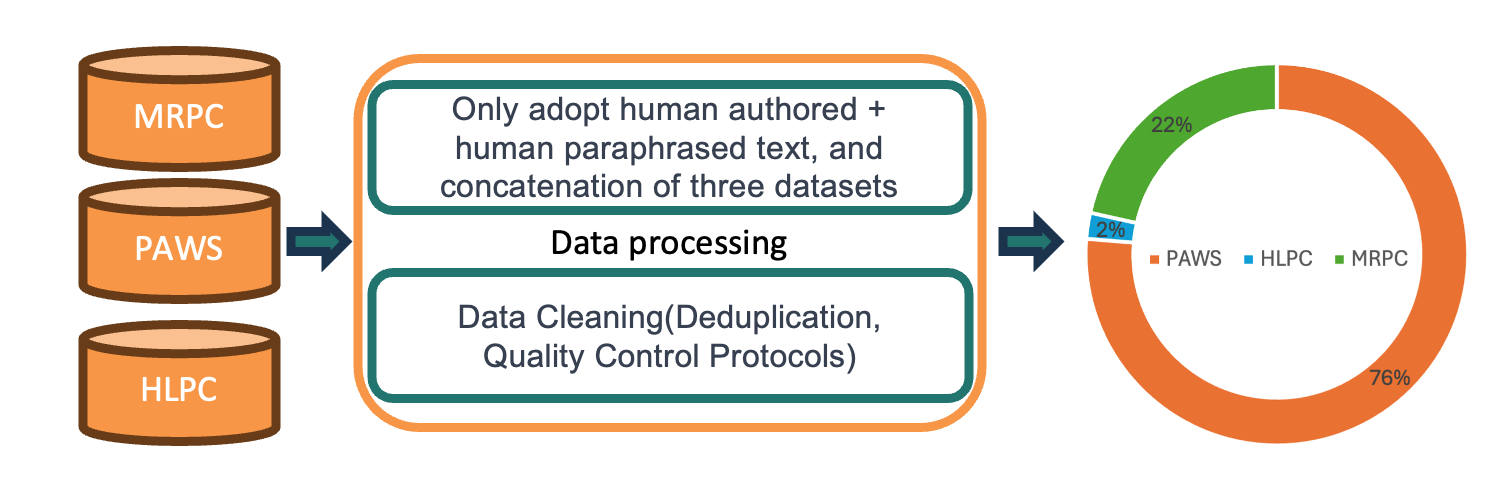}
    \caption{The complete integration of HLPC, MRPC, and PAWS datasets follows a systematic pipeline that encompasses data loading, standardization, quality control, and deduplication. This comprehensive approach ensures data integrity while maximizing the utility of each source dataset..}
    \label{fig:preprocessing}
\end{figure*}
The experiment utilizes 16,233 human-authored sentences sourced from three established datasets and followed by pipeline as showed in Figure \ref{fig:preprocessing}:
\begin{itemize}
    \item \textbf{MRPC} (Microsoft Research Paraphrase Corpus) \cite{dolan2005automatically}
    \item \textbf{HLPC} (Human-Like Paraphrase Corpus)\cite{lau2024understandingeffectshumanwrittenparaphrases}
    \item \textbf{PAWS} (Paraphrase Adversaries from Word Scrambling)\cite{paws2019naacl}
\end{itemize}

\subsection{MRPC processing}
The Microsoft Research Paraphrase Corpus (MRPC) \cite{dolan2005automatically} is a widely-used benchmark dataset for paraphrase detection, containing sentence pairs extracted from online news sources with human annotations indicating semantic equivalence. For our dataset construction, we extract only verified paraphrase pairs (\texttt{label == 1}), utilizing \texttt{sentence1} as human original text and \texttt{sentence2} as human paraphrased text. This filtering ensures high-quality semantic equivalence relationships while maintaining the news domain characteristics.

\subsection{PAWS processing}
The Paraphrase Adversaries from Word Scrambling (PAWS) dataset \cite{paws2019naacl} is specifically designed to challenge paraphrase identification systems with adversarial examples. The dataset contains sentence pairs derived from Wikipedia and Quora, where paraphrases are created through controlled word scrambling and substitution techniques, making them particularly challenging for automated detection systems while maintaining semantic equivalence. In our dataset construction, we only adopted the PAWS-QQP version where it adopted source data from QQP \cite{wang2017bilateralmultiperspectivematchingnatural}.We utilize the \texttt{labeled\_final} subset and extract only verified paraphrase pairs (\texttt{label == 1}), treating \texttt{sentence1} as human original text and \texttt{sentence2} as human paraphrased text. This approach ensures we capture the challenging paraphrase relationships that PAWS is designed to represent.

\subsection{HLPC processing}
The Human \& LLM Paraphrase Collection (HLPC) \cite{lau2024understandingeffectshumanwrittenparaphrases} is a comprehensive dataset that aggregates paraphrase data from multiple established sources including MRPC, XSum, QQP, and Multi-PIT. The dataset contains both human-authored paraphrases and machine-generated paraphrases produced by various language models (BART, DIPPER), providing a rich resource for studying different paraphrasing approaches and their characteristics.
However, we think the generated variants of HLPC is outdated since it mainly uses GPT-2-XL as main language model. Hence, we only utilize \texttt{originalSentence1} and \texttt{originalSentence2} to extract high-quality human paraphrase pairs, ensuring consistency with human annotation standards while leveraging the multi-source diversity of the collection.

\subsection{Preprocessing}
Given the potential overlap between datasets (particularly between HLPC and MRPC, as HLPC incorporates MRPC data), we implement a systematic deduplication process to prevent data leakage. Meanwhile, to ensure the data quality, we have strict data quality protocols on preprocessing. 

\paragraph{Quality Control Protocols}
Beyond deduplication, we implement comprehensive quality control measures:

\begin{enumerate}
    \item \textbf{Text Length Validation}: Remove entries with texts shorter than 10 characters or longer than 1000 characters
    \item \textbf{Encoding Validation}: Ensure proper UTF-8 encoding and remove entries with encoding issues
\end{enumerate}

\paragraph{Similarity-Based Duplicate Detection}

We employ TF-IDF vectorization combined with cosine similarity to identify near-duplicate content across the combined dataset:

\begin{equation}
\text{Similarity}(t_i, t_j) = \frac{\mathbf{v}_i \cdot \mathbf{v}_j}{|\mathbf{v}_i||\mathbf{v}_j|}
\end{equation}

where $\mathbf{v}_i$ and $\mathbf{v}_j$ are TF-IDF vectors for texts $t_i$ and $t_j$.

\begin{algorithm}[H]
\caption{Deduplication Process}
\label{algo:deduplication}
\begin{algorithmic}[1]
\STATE Compute TF-IDF vectors for all \texttt{human\_original\_text} entries
\STATE Calculate pairwise cosine similarities
\FOR{each text pair $(t_i, t_j)$ where $\text{Similarity}(t_i, t_j) > \theta$}
    \STATE Identify as potential duplicate
    \STATE Retain entry with higher dataset priority: PAWS > MRPC > HLPC
    \STATE Mark duplicate for removal
\ENDFOR
\STATE Remove identified duplicates from combined dataset
\end{algorithmic}
\end{algorithm}

When applying algorithm \ref{algo:deduplication} to remove the duplication in concatenated dataset, we set the threshold $\theta$ to be 0.85 to ensure all adopted human-authored texts are unique.

\section{Intrinsic Mechanisms of Paraphrase Attacks}

\subsection{Experiment 1: Semantic Equivalence versus Paraphrasing in Representation Space}
\label{app:experiment1}
\subsubsection{Research Objective}
In this experiment, our primary goal is to verify whether Hypothesis 1 holds: that paraphrasing induces a distinct semantic transformation, differing from text generated via ``semantic equivalence'' prompting. We assess this by comparing the embedding spaces of LLM-paraphrased and LLM-generated texts.

\subsubsection{Data Preparation}

We collected human-authored original sentences from three established paraphrase datasets, following the preprocessing pipeline shown in Figure \ref{fig:preprocessing}

After applying quality control and deduplication procedures (detailed in Appendix A), we obtained 16,233 unique human-authored sentences that serve as the foundation for generating the other two text categories.

\paragraph{LLM-Generated Text Creation}
\label{app:experiment1_llm_generated}
Using the 16,233 human-authored sentences as source material, we generated semantically equivalent sentences through LLM prompting. Unlike paraphrasing, this generation process aims to preserve the original meaning and structure without explicit rewording. We employed the following semantic equivalence prompt:

\begin{quote}
\small
\texttt{Given the following sentence, generate a new sentence that is semantically equivalent, preserving the original meaning and structure as closely as possible. Do not paraphrase or reword unnecessarily.}

\texttt{\{text\}}

\texttt{Generated sentence:}
\end{quote}

This approach produces LLM-generated text that maintains close semantic alignment with human-authored sources while exhibiting characteristic LLM generation patterns.

\paragraph{LLM-Paraphrased Text Creation}
\label{app:experiment1_paraphrased}
From the same 16,233 human-authored sentences, we generated paraphrased versions using explicit paraphrasing instructions. This category represents intentional lexical and syntactic transformation while preserving semantic content. The paraphrasing prompt was:

\begin{quote}
\small
\texttt{Please paraphrase the following text while maintaining its original meaning:}

\texttt{\{text\}}

\texttt{Paraphrased text:}
\end{quote}

This systematic approach yields three parallel text categories—human-authored, LLM-generated, and LLM-paraphrased—each containing 16,233 sentences, enabling controlled comparative analysis of semantic representations across text origins.

\subsubsection{Experimental Significance}
The experimental framework addresses two critical hypotheses:
\begin{itemize}
    \item \textbf{H1}: If LLM Generated $\approx$ LLM Paraphrased semantically, then paraphrase attacks exploit the same semantic space as original generation
    \item \textbf{H2}: If LLM Generated $\neq$ LLM Paraphrased, paraphrases create a distinct "attack space" requiring separate detection strategies
\end{itemize}

\subsubsection{Embedding Generation}

\noindent\textbf{BGE-M3 Model Configuration}
We employ the BGE-M3 (BAAI General Embedding Model) for generating high-dimensional semantic representations. The whole embedding generation process is illustrated in Algorithm.\ref{alg:embedding_generation}.

\begin{algorithm}[H]
\caption{Embedding Generation Process}
\label{alg:embedding_generation}
\begin{algorithmic}[1]
\STATE Initialize OpenAI client with BGE-M3 endpoint
\STATE \textbf{Input:} Text corpus $T = \{T_{\text{human}}, T_{\text{LLM}}, T_{\text{para}}\}$
\FOR{each text category\\ $t \in \{\text{human}, \text{LLM}, \text{para}\}$}
    \FOR{each sentence $s \in T_t$}
        \STATE $e_s \leftarrow \text{BGE-M3}(s)$ \COMMENT{Generate embedding vector}
    \ENDFOR
    \STATE Store embeddings as $E_t = \{e_s : s \in T_t\}$
\ENDFOR
\STATE \textbf{Output:} Embedding matrices \\ $\{E_{\text{human}}, E_{\text{LLM}}, E_{\text{para}}\}$
\end{algorithmic}
\end{algorithm}

\subsubsection{Distance Analysis}
\begin{figure*}[t]
    \centering
     \includegraphics[width=\textwidth]{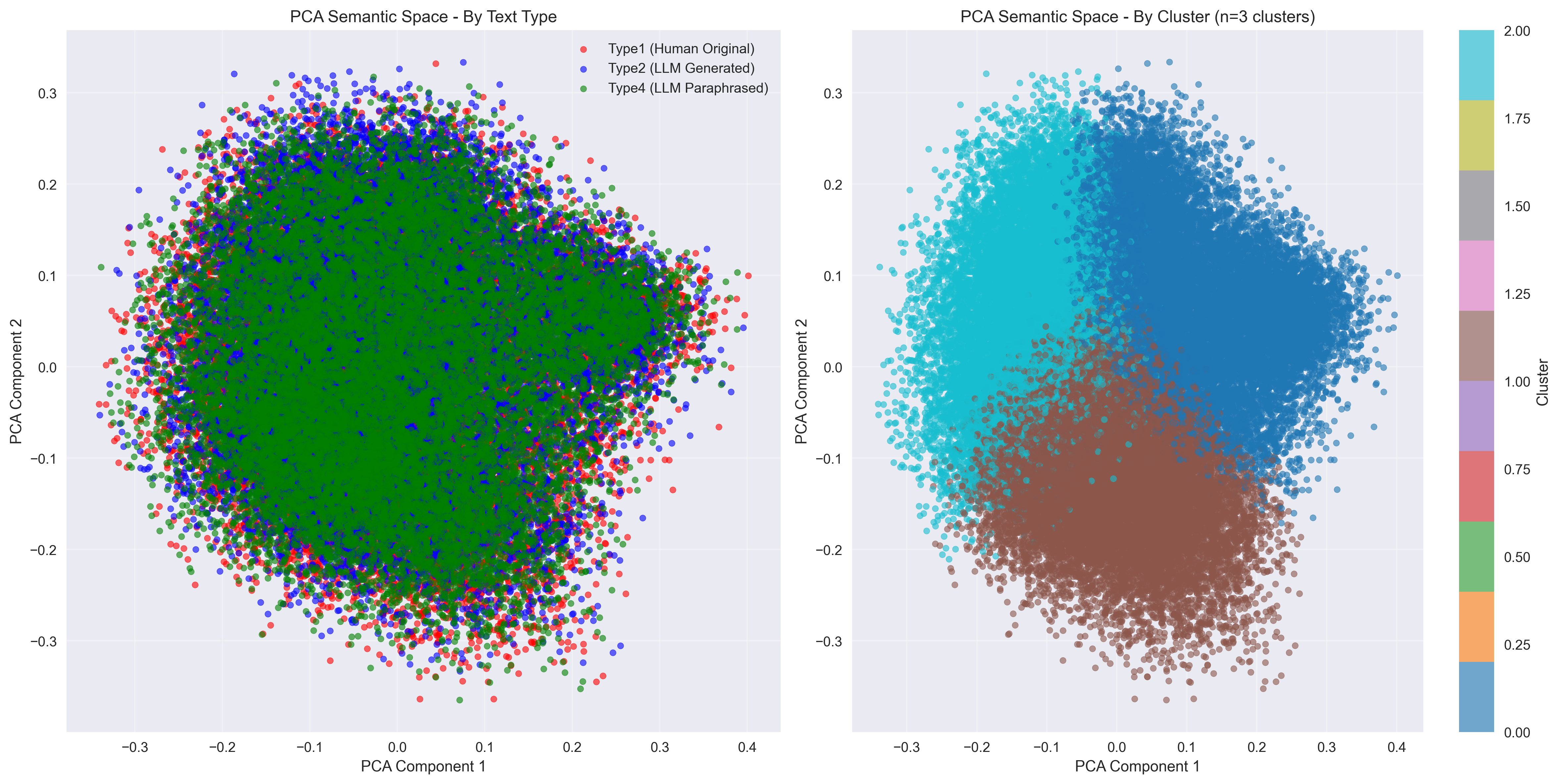}
    \caption{PCA projection of semantic space (left) and K-means clustering results (right, $k=3$). Despite measurable distance differences (Table~\ref{tab:pairwise_distances}), text categories show substantial overlap in 2D projection, indicating that distinguishing information exists in higher dimensions beyond principal components.}
    \label{fig:pca_clustering}
\end{figure*}
\noindent\textbf{Distance Metrics}
We compute three complementary distance metrics between embedding pairs to capture different aspects of semantic similarity:

\paragraph{cosine similarity}
Measures angular similarity between embedding vectors:
\begin{equation}
\label{eq:cosine_similarity}
\small
d_{\text{cosine}}(\mathbf{u}, \mathbf{v}) = 1 - \frac{\mathbf{u} \cdot \mathbf{v}}{|\mathbf{u}||\mathbf{v}|} = 1 - \frac{\sum_{i=1}^{n} u_i v_i}{\sqrt{\sum_{i=1}^{n} u_i^2} \sqrt{\sum_{i=1}^{n} v_i^2}}
\end{equation}

where $\mathbf{u}, \mathbf{v} \in \mathbb{R}^n$ are embedding vectors. Range: $[0, 1]$ where 0 indicates identical direction and 1 indicates orthogonality.

\paragraph{Euclidean Distance}
Computes straight-line distance in embedding space:
\begin{equation}
\label{eq:euclidean_distance}
\small
d_{\text{euclidean}}(\mathbf{u}, \mathbf{v}) = \sqrt{\sum_{i=1}^{n} (u_i - v_i)^2} = |\mathbf{u} - \mathbf{v}|_2
\end{equation}

\paragraph{Manhattan Distance}
Calculates city-block distance:
\begin{equation}
\label{eq:manhattan_distance}
\small
d_{\text{manhattan}}(\mathbf{u}, \mathbf{v}) = \sum_{i=1}^{n} |u_i - v_i| = |\mathbf{u} - \mathbf{v}|_1
\end{equation}

\paragraph{Pairwise Distance Computation}
For each distance metric $d$, we compute average distances between text type pairs:
\begin{align}
\label{eq:pairwise_distance}
\small
D_{H,L}^{(d)} &= \frac{1}{|E_H| \cdot |E_L|} \sum_{e_h \in E_H} \sum_{e_l \in E_L} d(e_h, e_l) \\
D_{H,P}^{(d)} &= \frac{1}{|E_H| \cdot |E_P|} \sum_{e_h \in E_H} \sum_{e_p \in E_P} d(e_h, e_p) \\
D_{L,P}^{(d)} &= \frac{1}{|E_L| \cdot |E_P|} \sum_{e_l \in E_L} \sum_{e_p \in E_P} d(e_l, e_p)
\end{align}
where $H$, $L$, and $P$ denote human-authored, LLM-generated, and paraphrased text, respectively.

\subsubsection{Semantic Space Exploration}
\label{appB:experiment1_semantic_space}
\begin{figure}[t]
    \centering
    \includegraphics[width=\columnwidth]{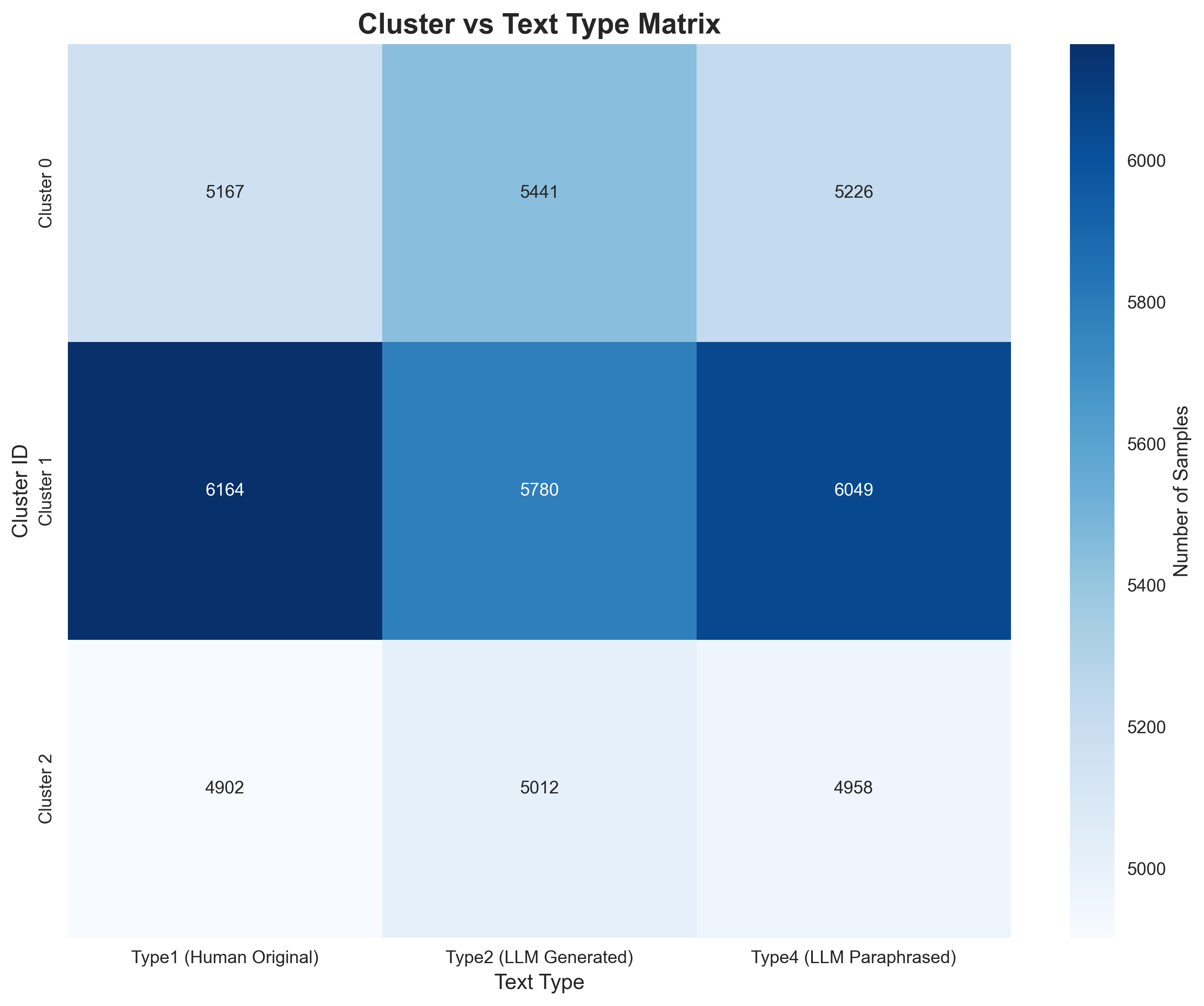}
    \caption{Clustering label distribution across human-authored, LLM-generated, and LLM-paraphrased text. The matrix reveals that low-dimensional representation space makes it difficult to distinguish between the three text types.}
    \label{fig:cluster_type_matrix}
\end{figure}
\paragraph{Dimensionality Reduction via PCA}
We apply Principal Component Analysis (PCA) to project the high-dimensional BGE-M3 embeddings (1024 dimensions) into 2D visualization space. The PCA transformation preserves the directions of maximum variance, enabling clear visualization of the primary semantic relationships between the three text categories in the combined embedding space $E_{\text{combined}} = [E_{\text{human}}; E_{\text{generated}}; E_{\text{paraphrased}}]$. Figure \ref{fig:pca_clustering} shows the PCA visualization. 

\paragraph{Unsupervised clustering via KMeans}
We apply K-Means clustering with $k=3$ clusters to the combined embedding space to identify natural semantic groupings. The algorithm partitions the embeddings into three clusters by minimizing within-cluster sum of squares, with random initialization and iterative optimization until convergence. Figure \ref{fig:pca_clustering} right graph shows the KMeans clustering visualization, and Figure \ref{fig:cluster_type_matrix} shows the detailed label distribution. 

\subsection{Experiment 2: Iterative Paraphrasing in Representation Space}
\label{App:Experiment2}
\subsubsection{Research Objective}
This experiment investigates \textbf{Hypothesis 2}: Iterative paraphrasing makes text become more coherent, deviating from common LLM-generated patterns and moving closer to human-authored texts. We analyze semantic drift through multiple iterations of paraphrasing to understand how text evolves in semantic space over successive transformations.

\subsubsection{Data Preparation}
The experiment randomly samples 100 human authored texts from the combined dataset we processed(Detailed in Appendix. \ref{Appendix_A}. As for the iterative paraphrasing, we 

\subsubsection{Experiment Procedure}

\paragraph{Representation Space Choice}
For each iteration, we extract two complementary representations:

\begin{enumerate}
    \item \textbf{Hidden States}: Last-layer hidden states from Qwen3-4B-Instruct with mean pooling across sequence length
    \item \textbf{Semantic Embeddings}: BGE-M3 embeddings via Novita AI API for semantic space analysis
\end{enumerate}

\paragraph{Distance Analysis}
We conduct two complementary distance analyses to examine semantic drift under iterative paraphrasing:

\textbf{Analysis 1: Progressive Semantic Drift.} We measure how each paraphrasing iteration affects semantic distance by comparing consecutive iterations. Specifically, we compute distances between iteration $i$ and iteration $i+1$ for both human-authored and LLM-generated text that have undergone 1-10 paraphrasing iterations. This analysis reveals the incremental semantic changes introduced by each successive paraphrasing step. The result table can be found in Table. \ref{tab:iterative_distances}.

\textbf{Analysis 2: Cross-Category Semantic Distance.} We measure semantic distances between original human-authored text and paraphrased human-authored text (iterations 1-10), as well as between original LLM-generated text and paraphrased human-authored text (iterations 1-10). This analysis examines how iterative paraphrasing of human text affects its semantic proximity to both human-authored and LLM-generated references, revealing potential convergence or divergence patterns across text categories.The result table can be found in Table. \ref{tab:semantic_distances_detailed}.

\begin{table}[t]
\centering
\small
\caption{Cosine similarity: 1.single iteration of paraphrased human-authored text versus 2-10 iterations of paraphrased human-authored text 2.single iteration of paraphrased llm-generated text versus 2-10 iterations of paraphrased llm-generated text based on BGE-m3 embedding and Qwen3-4B final hidden state. H. is indicating the human-authored text, and L. is indicating the LLM-generated text.}
\label{tab:iterative_distances}
\begin{tabular}{@{}p{1cm}lccccc@{}}
\toprule
\multirow{2}{*}{\textbf{Repr.}} & \multirow{2}{*}{\textbf{Type}} & \multicolumn{5}{c}{\textbf{Iteration}} \\
\cmidrule(lr){3-7}
& & \textbf{2} & \textbf{4} & \textbf{6} & \textbf{8} & \textbf{10} \\
\midrule
\multirow{2}{*}{\shortstack[l]{BGE\\Emb.}} 
& H & 0.048 & 0.072 & 0.083 & 0.092 & 0.100 \\
& L & 0.047 & 0.068 & 0.077 & 0.087 & 0.091 \\
\midrule
\multirow{2}{*}{\shortstack[l]{Hid.\\Stat.}} 
& H & 0.012 & 0.022 & 0.015 & 0.017 & 0.019 \\
& L & 0.011 & 0.014 & 0.015 & 0.017 & 0.018 \\
\bottomrule
\end{tabular}
\end{table}

\begin{table}[t]
\centering
\small
\caption{Semantic distance between two text types: (1) human-authored text versus 1-10 iterations of paraphrased human-authored text (2) LLM-generated text versus 1-10 iterations of paraphrased human-authored text. }
\label{tab:semantic_distances_detailed}
\begin{tabular}{@{}llcc@{}}
\toprule
\multirow{2}{*}{\textbf{Text Type}} & \multirow{2}{*}{\textbf{Iter.}} & \textbf{Cosine} & \textbf{Euclidean} \\
& & \textbf{Distance} & \textbf{Distance} \\
\midrule
\multirow{10}{*}{\shortstack[l]{Human-\\Authored}} 
& 1 & 0.064 & 0.342 \\
& 2 & 0.085 & 0.394 \\
& 3 & 0.099 & 0.426 \\
& 4 & 0.107 & 0.443 \\
& 5 & 0.118 & 0.465 \\
& 6 & 0.122 & 0.472 \\
& 7 & 0.125 & 0.478 \\
& 8 & 0.128 & 0.484 \\
& 9 & 0.129 & 0.486 \\
& 10 & 0.134 & 0.494 \\
\midrule
\multirow{10}{*}{\shortstack[l]{LLM-\\Generated}} 
& 1 & 0.700 & 1.182 \\
& 2 & 0.698 & 1.180 \\
& 3 & 0.696 & 1.179 \\
& 4 & 0.697 & 1.180 \\
& 5 & 0.696 & 1.179 \\
& 6 & 0.697 & 1.179 \\
& 7 & 0.697 & 1.180 \\
& 8 & 0.699 & 1.181 \\
& 9 & 0.697 & 1.179 \\
& 10 & 0.698 & 1.180 \\
\bottomrule
\end{tabular}
\end{table}
For both analyses, we employ three complementary distance metrics to capture different aspects of semantic dissimilarity: cosine similarity, Euclidean distance, and Manhattan distance (see Equations \ref{eq:cosine_similarity}, \ref{eq:euclidean_distance}, and \ref{eq:manhattan_distance}).

Analysis 1 uses centroid-based distance to measure population-level drift. For each iteration $i$ and text type $t$, we compute population centroids:
\begin{equation}
\mathbf{c}_{i,t} = \frac{1}{N} \sum_{j=1}^{N} \mathbf{e}_{i,t,j}
\end{equation}
where $\mathbf{e}_{i,t,j}$ represents the embedding of sample $j$ at iteration $i$ for text type $t$. Sequential distance analysis then tracks semantic drift between consecutive iterations:
\begin{equation}
\Delta d_{i \rightarrow i+1} = d(\mathbf{c}_{i,t}, \mathbf{c}_{i+1,t})
\end{equation}

Analysis 2 employs the pairwise distance calculation described in Equation \ref{eq:pairwise_distance}, measuring distances between the reference texts (human-authored original and LLM-generated original) and iteratively paraphrased human-authored text at each iteration level.

\paragraph{PCA Trajectory Analysis}
\begin{figure*}[t]
    \centering
    \includegraphics[width=\textwidth]{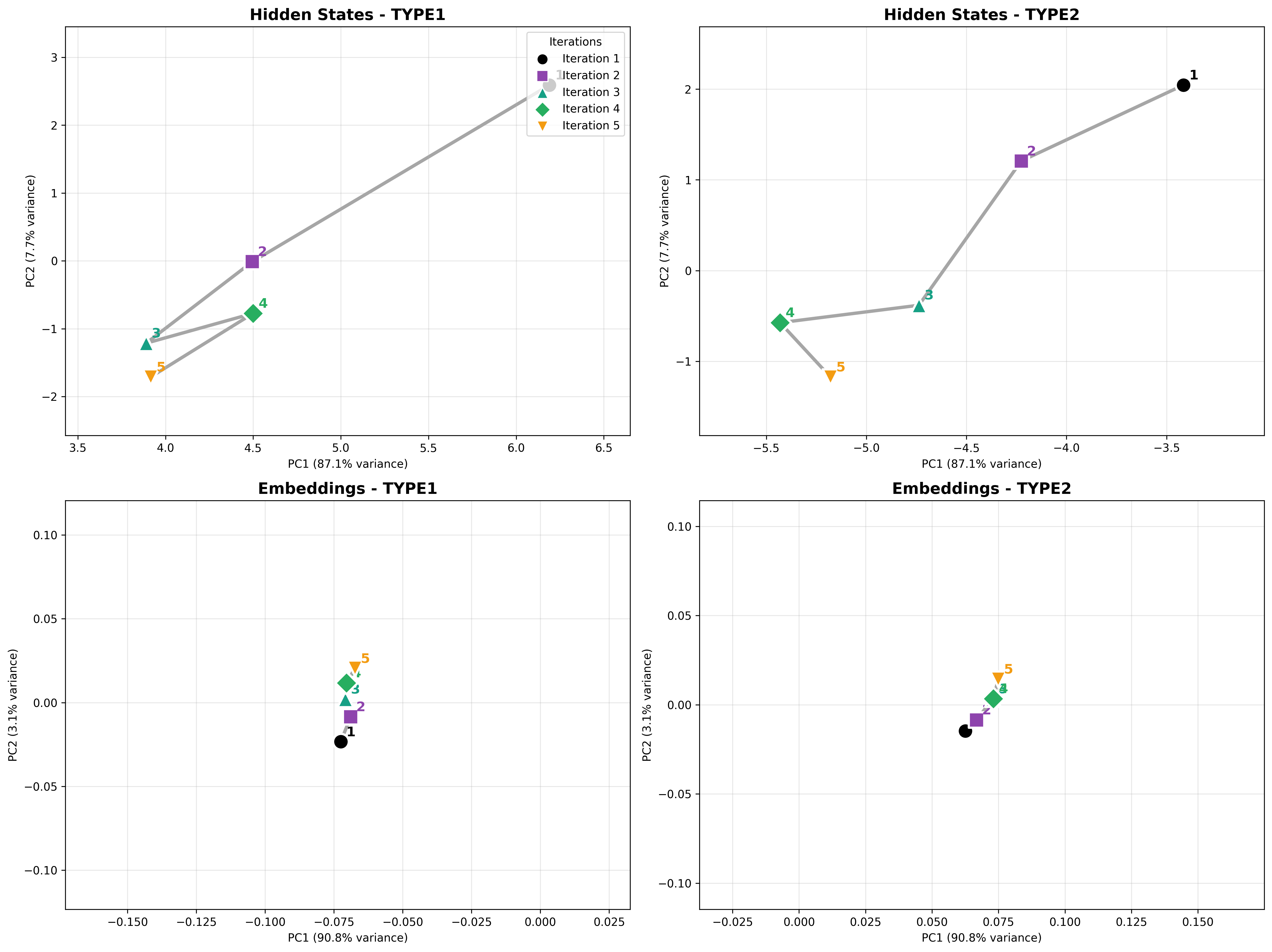}
    \caption{The 5-iteration Centroid trajectories under PCA(n=2). Top: Hidden state space (left: human-origin, right: LLM-origin). Bottom: Embedding space. }
    \label{fig:trajectories_5itr}
\end{figure*}

\begin{figure*}[t]
    \centering
    \includegraphics[width=\textwidth]{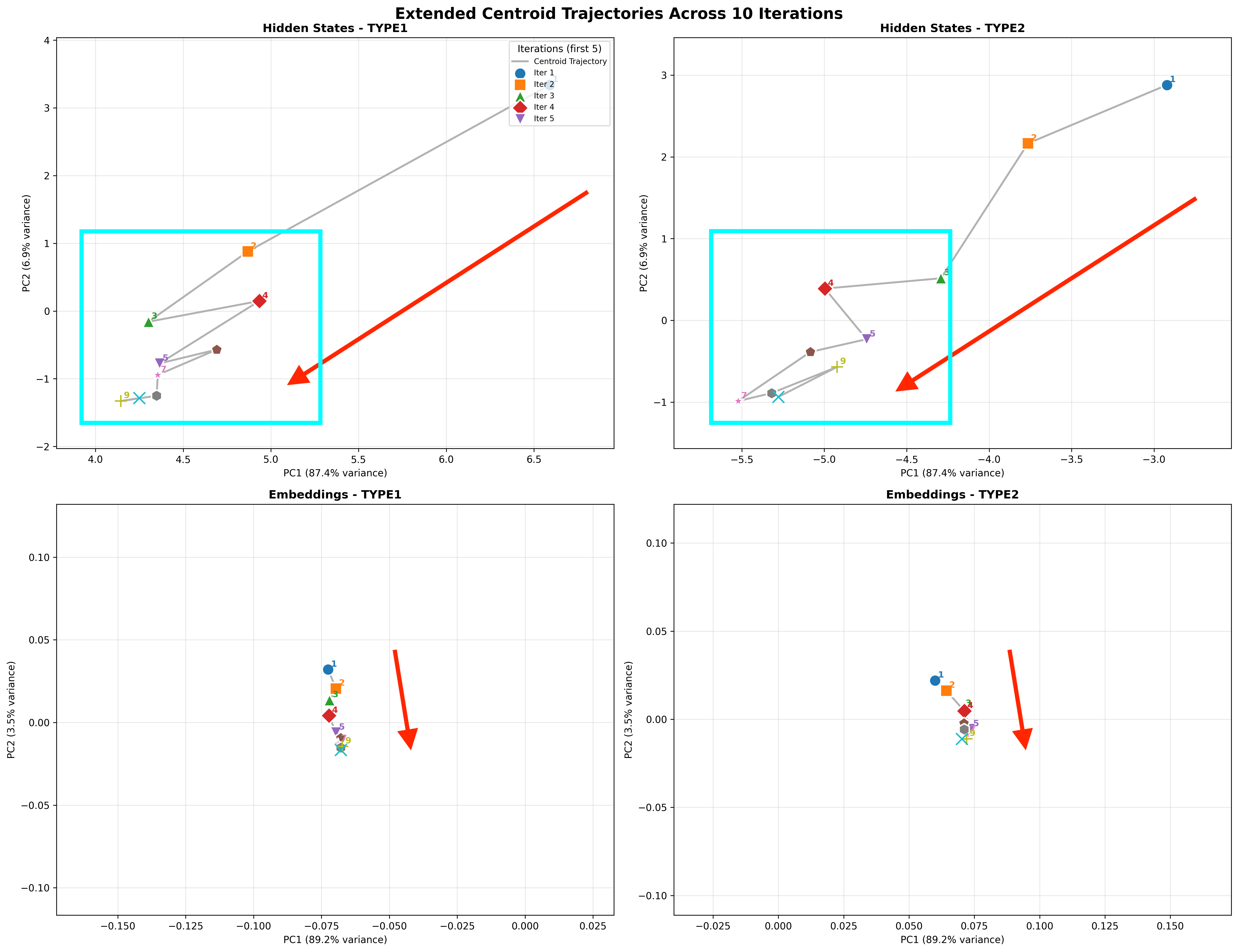}
    \caption{Extended 10-iteration centroid trajectories showing semantic drift patterns. Top: Hidden state space (left: human-origin, right: LLM-origin). Bottom: Embedding space. Both origins move in parallel directions, with trajectories converging toward overlapping regions in later iterations.}
    \label{fig:trajectories_10itr}
\end{figure*}
To visualize semantic drift patterns across iterations, we apply Principal Component Analysis (PCA) to project high-dimensional embeddings into 2D visualization space. This trajectory analysis tracks how text representations evolve through successive paraphrasing iterations.

We initially conducted a 5-iteration analysis to identify semantic drift patterns. However, the resulting trajectories did not reveal sufficiently clear patterns to draw robust conclusions about long-term semantic behavior. The result figure can be found in Figure.\ref{fig:trajectories_5itr}. Consequently, we extended the analysis to 10 iterations, which provided more definitive evidence of semantic drift trajectories and convergence patterns.

The PCA trajectory analysis follows the procedure outlined in Algorithm \ref{alg:pca_trajectory}:

\begin{algorithm}[H]
\caption{PCA Trajectory Analysis}
\label{alg:pca_trajectory}
\begin{algorithmic}[1]
\STATE Collect all embeddings across iterations: $E = \{\mathbf{e}_{i,t,j} : i \in [1,N_{iter}], t \in \{\text{type1}, \text{type2}\}, j \in [1,N_{samples}]\}$
\STATE Standardize features: $\tilde{E} = \text{StandardScaler}(E)$
\STATE Apply PCA: $E_{\text{PCA}} = \text{PCA}_{n=2}(\tilde{E})$
\STATE Compute iteration centroids in PCA space: $\mathbf{c}_{i,t}^{\text{PCA}} = \frac{1}{N} \sum_{j=1}^{N} \mathbf{e}_{i,t,j}^{\text{PCA}}$
\STATE Track centroid trajectories: $\mathcal{T}_t = \{\mathbf{c}_{i,t}^{\text{PCA}} : i \in [1,N_{iter}]\}$
\end{algorithmic}
\end{algorithm}

To quantify the magnitude of semantic drift, we compute the total Euclidean displacement of centroids across iterations:
\begin{equation}
\text{Total Drift}_t = \sum_{i=1}^{N_{iter}-1} ||\mathbf{c}_{i+1,t}^{\text{PCA}} - \mathbf{c}_{i,t}^{\text{PCA}}||_2
\end{equation}

Figure \ref{fig:trajectories_10itr} visualizes the resulting trajectories in PCA space, where each point represents the centroid of all samples at a given iteration, and connecting lines trace the semantic evolution path. The trajectories reveal the potential two insights: Universal Directional Drift and Intermediate Laundering Region, which detailed demonstrated in Section.\ref{finding_results}.

\section{Detailed Methodology}
\label{Appendix_C}
\subsection{Dataset Preparation}
Our benchmark builds upon the human-authored sentences (Type 1) and 
human paraphrases (Type 3) from three established datasets: MRPC, HLPC, 
and PAWS. The detailed preprocessing pipeline, including deduplication 
and quality control procedures, is described in Appendix~\ref{Appendix_A}. 
This foundation provides 16,233 unique human-authored sentences and human-paraphrased human texts, from which we systematically generate the remaining text types (Type 2, 4, 5) using the pipeline described in Appendix.\ref{appC:data_generation}.
\subsection{Text Type Taxonomy}
\label{appC:taxonomy_definition}
We establish a five-category taxonomy to systematically analyze different text generation and paraphrasing patterns:

\begin{itemize}
    \item \textbf{Type 1}: Human original text -- authentic human-authored sentences
    \item \textbf{Type 2}: LLM-generated text -- synthetically generated content maintaining semantic equivalence to Type 1(generated by sentence completion method)
    \item \textbf{Type 3}: Human-paraphrased human original text -- human-authored paraphrases of Type 1 sentences
    \item \textbf{Type 4}: LLM-paraphrased human original text -- machine-generated paraphrases of Type 1 sentences
    \item \textbf{Type 5}: LLM-iteratively-paraphrased LLM-generated text -- machine-generated paraphrases of Type 2 sentences with multiple iteration levels(using the same paraphrasing prompting as type4)
\end{itemize}

\subsection{Data Generation Pipeline}
\label{appC:data_generation}
\subsubsection{Technical Architecture}
Our data generation system employs a modular, configuration-driven architecture with model selection optimized for each text type's specific requirements. This approach ensures high-quality generation while leveraging the strengths of different specialized models.The use of multiple models for paraphrasing (Type 4 and 5) is a deliberate choice to create a diverse dataset that is not biased toward the stylistic quirks of a single paraphraser, thereby presenting a more realistic and challenging test for detectors.

The pipeline implements three sequential generation modules:

\begin{enumerate}
    \item \textbf{Type 2 Generation Module}: Sentence completion-based text synthesis using \textbf{Google Gemini-2.5-Pro}
    \item \textbf{Type 4 Generation Module}: multiple model-used paraphrasing combining DIPPER paraphraser \cite{krishna2023paraphrasingevadesdetectorsaigenerated}, Gemini-2.5-Pro with prompt-based instructions, and LLaMA-3-8B paraphrase fine-tuned models \cite{mradermacher2024llama31paraphrase}
    \item \textbf{Type 5 Generation Module}: Iterative paraphrasing using the same multi-model approach as Type 4
\end{enumerate}

\subsubsection{Type 2 Generation: Sentence Completion Method}

For Type 2 text generation, we implement a \textbf{sentence completion approach} designed to produce text that is contextually grounded in the original human sentence while allowing for natural, unconstrained continuation. This mirrors how a user might leverage an LLM for co-writing or content expansion. The process involves:

\begin{enumerate}
    \item \textbf{Keyword Extraction}: Using SpaCy's named entity recognition and dependency parsing to identify salient keywords from Type 1 sentences
    \item \textbf{Prefix Extraction}: Extracting the first 20\% of tokens from the original sentence as contextual seed
    \item \textbf{Length Constraints}: Computing target length parameters with $\pm$20\% tolerance
\end{enumerate}

\noindent\textbf{Generation Prompt Template}
\begin{quote}
\small
\texttt{Continue this text naturally and coherently:}

\texttt{"\{sentence\_prefix\}"}

\texttt{Requirements:}
\begin{itemize}[nosep,leftmargin=1em]
    \item \texttt{Target length: \textasciitilde\{target\_length\} characters total}
    \item \texttt{Maximum length: \{max\_length\} characters}
    \item \texttt{Keywords to include: \{keywords\}}
    \item \texttt{Write in a natural, fluent style}
\end{itemize}

\texttt{Return ONLY the completed text with no labels, quotes, explanations, or alternatives.}

\texttt{Completion:}
\end{quote}

\subsubsection{Type 4 Generation: Direct Paraphrasing}
Type 4 generation, which simulates a direct attempt to launder human-written content, employs \textbf{prompt-based paraphrasing} using carefully engineered instructions. This approach prioritizes semantic preservation while encouraging significant lexical and syntactic variation. The length tolerance is set to ±30\% to accommodate natural paraphrasing variation.

\noindent\textbf{Paraphrasing Prompt Template}
\begin{quote}
\small
\texttt{Please paraphrase the following text while maintaining its original meaning:}

\texttt{\{text\}}

\texttt{Paraphrased text:}
\end{quote}

\subsubsection{Type 5 Generation: Iterative Paraphrasing}
To simulate more sophisticated evasion attempts where AI text is laundered multiple times, the Type 5 module implements \textbf{multi-iteration paraphrasing} of Type 2 texts. We support two levels: 1 and 3 iterations, where each iteration applies the paraphrasing prompt to the output of the previous one.

\textbf{Iteration Control Mechanisms}:
\begin{itemize}
    \item \textbf{Temperature Scaling}: Base temperature (0.8) increases by 0.1--0.15 per iteration level to enhance diversity
    \item \textbf{Convergence Detection}: Automatic termination when consecutive iterations achieve $>$95\% similarity
    \item \textbf{Length Tolerance}: Expanded to $\pm$40\% to accommodate cumulative variation across iterations
\end{itemize}

\subsection{Data Quality Assessment}
\label{appC:data_quality}

To ensure the integrity and characteristics of our generated dataset, we employ three complementary quality metrics: \textbf{Jaccard similarity}, \textbf{perplexity}, and \textbf{self-BLEU} scores.

\textbf{Jaccard Similarity Analysis}: The inter-type similarity matrix reveals expected semantic relationships across text types. Human original text (Type 1) demonstrates highest similarity with human paraphrases (Type 3, Jaccard = 0.798), confirming semantic preservation in human paraphrasing. LLM-generated text (Type 2) shows moderate cross-similarity with other synthetic types, indicating consistent generation patterns. Notably, iterative paraphrasing exhibits controlled diversity: Type 5 first-iteration maintains reasonable similarity with its source (0.469 with Type 2), while third-iteration paraphrasing (Type 5-3) shows increased lexical divergence (0.423 with Type 2), demonstrating successful iterative transformation without complete semantic drift.

\begin{figure}[htbp]
    \centering
    \includegraphics[width=0.5\textwidth]{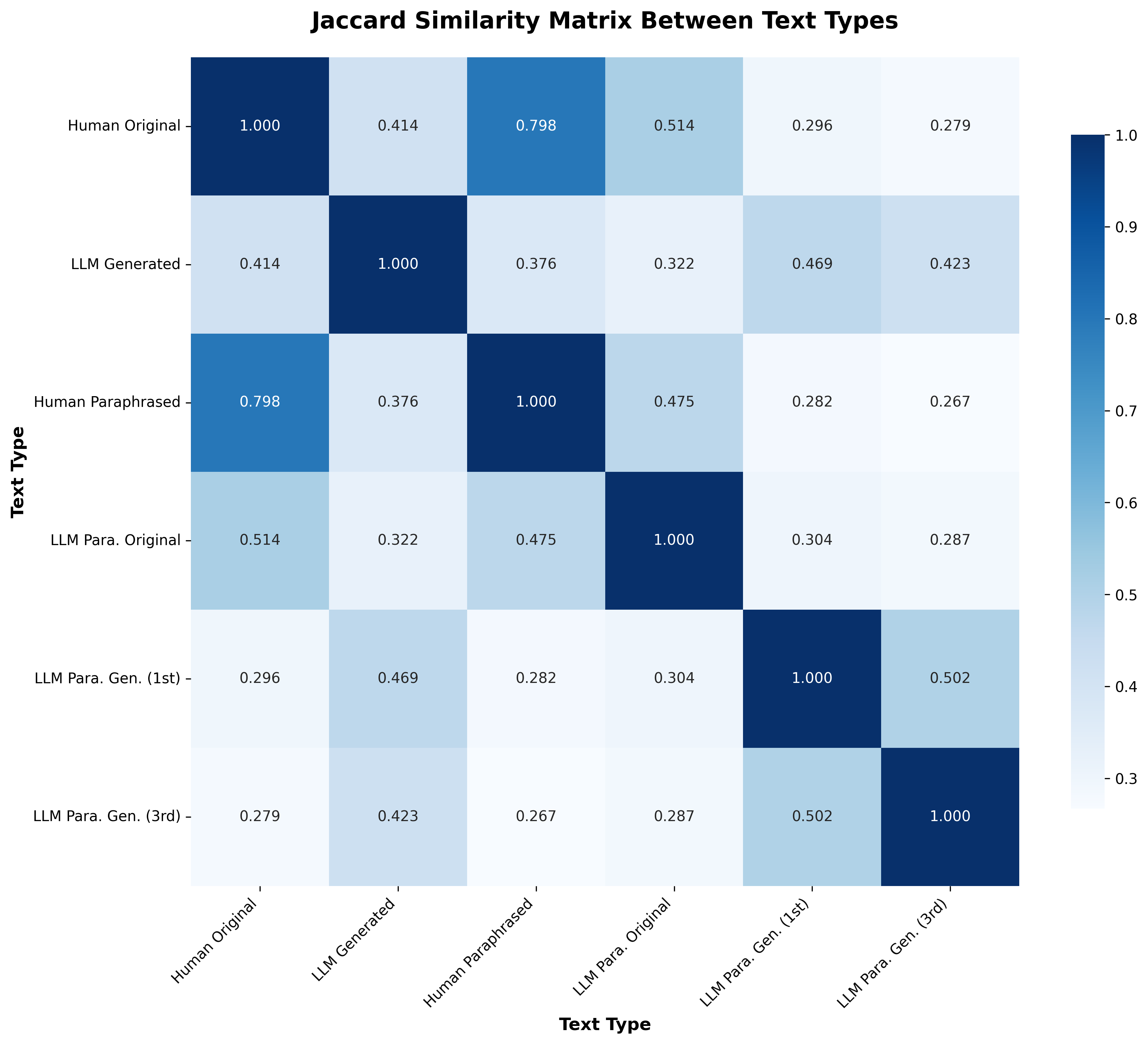}
    \caption{Jaccard similarity matrix between different text types.}
    \label{fig:jaccard_similarity}
\end{figure}

\textbf{Perplexity Evaluation}: We assess text predictability using perplexity scores computed with GPT-2-XL and LLaMA-2-7B as reference models. The quality metrics reveal consistent patterns across both evaluation models (Table~\ref{tab:quality_metrics}). 

Perplexity analysis demonstrates remarkable consistency between GPT-2-XL and LLaMA-2-7B evaluations. LLM-generated text (Type 2) exhibits the lowest perplexity scores across both models (GPT-2-XL: 77.84, LLaMA-2-7B: 42.61), indicating high predictability and suggesting that machine-generated content follows formulaic patterns readily recognized by different language model architectures. Human original and paraphrased texts (Types 1, 3) consistently demonstrate higher perplexity scores (GPT-2-XL: 106.78/107.47, LLaMA-2-7B: 49.57/49.45), suggesting greater linguistic variability and creativity that deviates from typical language model expectations. 

Notably, both reference models identify iterative paraphrasing as producing the most unpredictable content, with Type 5-3rd achieving the highest perplexity in GPT-2-XL (109.32) and among the highest in LLaMA-2-7B (50.23). This cross-model validation strengthens our conclusion that iterative paraphrasing successfully diversifies content away from conventional language model patterns. The consistently lower absolute perplexity values from LLaMA-2-7B (average: 46.46) compared to GPT-2-XL (average: 97.82) reflect architectural differences in predictive modeling, while maintaining similar relative rankings across text types.

\begin{table}[htbp]
\centering
\caption{Quality metrics across text types.}
\label{tab:quality_metrics}
\footnotesize
\begin{tabular}{lccc}
\hline
\textbf{Text Type} & \textbf{PPL-G2X} & \textbf{PPL-L7B} & \textbf{sBLEU} \\
\hline
Type 1 & 106.78 & 49.57 & 0.268 \\
Type 2 & 77.84 & 42.61 & 0.242 \\
Type 3 & 107.47 & 49.45 & 0.275 \\
Type 4 & 85.90 & 39.63 & 0.196 \\
Type 5-1st & 99.63 & 47.29 & 0.179 \\
Type 5-3rd & 109.32 & 50.23 & 0.170 \\
\hline
\textbf{Average} & \textbf{97.82} & \textbf{46.46} & \textbf{0.222} \\
\hline
\end{tabular}
\end{table}

\textbf{Self-BLEU Assessment}: Self-BLEU scores measure intra-type diversity within each text category, preventing over-generation of similar content. 

Self-BLEU scores showed in Table~\ref{tab:quality_metrics} demonstrate appropriate diversity levels across all text types. Human-authored content (Types 1, 3) shows moderate self-similarity (0.268, 0.275), while machine-processed texts exhibit progressively lower self-BLEU scores, with iterative paraphrasing achieving maximum diversity (Type 5-3rd: 0.170). This gradient confirms successful generation of varied content within each category.

\begin{table}[htbp]
\centering
\caption{Quality metrics comparison between PadBen and RAID datasets (average values).}
\label{tab:dataset_comparison}
\begin{tabular}{lccc}
\hline
\textbf{Dataset} & \textbf{Self-BLEU} & \textbf{PPL-G2X} & \textbf{PPL-L7B} \\
\hline
PadBen & 0.222 & 97.82 & 46.46 \\
RAID & 13.7 & 23.8 & 6.6 \\
\hline
\end{tabular}
\end{table}

\textbf{Comparison}: The comparison with the RAID \cite{raid} dataset reveals significant differences in quality metrics across multiple evaluation criteria (Table~\ref{tab:dataset_comparison}). PadBen exhibits substantially lower self-BLEU scores (0.222) compared to RAID (13.7), indicating approximately 62× higher intra-type diversity. This dramatic difference suggests that PadBen successfully generates more varied content within each text category, reducing the risk of repetitive patterns that could bias evaluation results.

Regarding perplexity evaluation, PadBen demonstrates consistently higher linguistic unpredictability across both reference models. Using GPT-2-XL, PadBen achieves 4.1× higher perplexity scores (97.82 vs 23.8), while LLaMA-2-7B evaluation shows an even more pronounced 7.0× difference (46.46 vs 6.61). This cross-model validation strengthens our findings, indicating that PadBen's generated content consistently presents more complex and diverse linguistic structures that deviate from conventional language model expectations regardless of the evaluation architecture.

The substantial perplexity differences across both evaluation models particularly benefit adversarial evaluation scenarios, as they suggest our synthetic text maintains sufficient complexity to challenge detection systems effectively. The consistency of these patterns across different model architectures (GPT-2-XL and LLaMA-2-7B) demonstrates that PadBen's quality advantages are not dependent on specific evaluation frameworks but represent genuine improvements in linguistic diversity and complexity.

\subsection{Dataset Statistics and Characteristics}
Our final dataset comprises \textbf{16232 sentence groups} across three source datasets, each containing all five text types. The systematic generation approach addresses previous limitations in existing datasets, particularly outdated model usage and inconsistent generation methodologies, providing a robust foundation for analyzing human versus machine text characteristics across multiple transformation levels.

\subsection{Detailed Task Introduction}
\label{appC:task_intro}

\subsubsection{Task 1: Paraphrase Source Attribution}

\textbf{Objective}: Evaluate detectors' ability to distinguish between human and machine paraphrasing without access to original source text.

\textbf{Data Configuration}: Utilize Type 3 (human-paraphrased) and Type 4 (LLM-paraphrased) texts as input. Human paraphrases are labeled as 0, while LLM-generated paraphrases receive label 1.

\textbf{Research Question}: Can AI detectors identify the authorship of paraphrased content when the original text is unavailable for comparison?

\textbf{Detection Challenge}: Without reference to original text, detectors must rely solely on intrinsic linguistic markers and stylistic patterns to differentiate human and machine paraphrasing strategies. This tests whether human and LLM paraphrasing exhibit distinguishable linguistic signatures.

\subsubsection{Task 2: General Text Authorship Detection}

\textbf{Objective}: Assess baseline detection performance on distinguishing original human-authored text from LLM-generated content.

\textbf{Data Configuration}: Utilize Type 1 (human original) and Type 2 (LLM-generated) texts as input. Human-authored content is labeled as 0, while LLM-generated text receives label 1. 

\textbf{Research Question}: How effectively can current detectors distinguish between authentic human writing and synthetically generated text?

\textbf{Detection Challenge}: This represents the foundational detection scenario that most existing systems are designed to address. Performance on this task establishes baseline capabilities and serves as a reference point for evaluating more complex detection scenarios.

\subsubsection{Task 3: AI Text Laundering Detection}

\textbf{Objective}: Evaluate detectors' resilience against AI text laundering through paraphrasing attacks.

\textbf{Data Configuration}: Utilize Type 4 (LLM-paraphrased human text) and Type 5-1st (single-iteration LLM-paraphrased LLM text) as input. LLM-paraphrased human content is labeled as 0, while laundered AI text receives label 1. 

\textbf{Research Question}: Can detectors identify the authorship of original content after one iteration of paraphrasing?

\textbf{Detection Challenge}: This task simulates a common evasion strategy where AI-generated content is paraphrased to mask its synthetic origin. The challenge lies in determining which text originated from human versus LLM sources before paraphrasing, where both texts have undergone identical machine transformation.

\subsubsection{Task 4: Iterative Paraphrase Depth Detection}

\textbf{Objective}: Assess detector ability to distinguish between shallow and deep iterative paraphrasing attacks.

\textbf{Data Configuration}: Utilize Type 5-1st (single-iteration paraphrased LLM text) and Type 5-3rd (triple-iteration paraphrased LLM text) as input. Shallow paraphrasing (1 iteration) is labeled as 0, while deep paraphrasing (3 iterations) receives label 1.

\textbf{Research Question}: Can detectors identify which text has undergone higher iteration of paraphrasing?

\textbf{Detection Challenge}: This represents a sophisticated evasion scenario where detectors must distinguish between different depths of iterative transformation applied to the same synthetic source. The task evaluates whether detection systems can identify progressive obfuscation levels.

\subsubsection{Task 5: Paraphrase attack Detection}

\textbf{Objective}: Evaluate detector resilience in the ultimate end-to-end evasion scenario, comparing original human writing against deeply laundered AI-generated text, mimicking the paraphrase attack scenario.

\textbf{Data Configuration}: Utilize Type 1 (human original) and Type 5-3rd (triple-iteration LLM-paraphrased LLM text) as input. Human text is labeled as 0, and deeply laundered AI text is labeled as 1.

\textbf{Research Question}: Does detector able to detect trace of AI authorship remain after iterative paraphrasing(paraphrase attack)?

\textbf{Detection Challenge}: This is the benchmark's final stress test, simulating the paraphrase attack. Success requires detectors to identify highly subtle, persistent machine-generation artifacts that have survived multiple layers of transformation, distinguishing deeply-laundered AI text from authentic human writing.

\section{Detailed Task Data Setup}
\label{appD:evaluation_setup}

\begin{algorithm}[H]
\caption{Single-Sentence Exhaustive Method}
\begin{algorithmic}[1]
\label{algo:exhaustive_method}
\STATE \textbf{Input:} Dataset $D$ with $n$ samples, Task specification $(T_A, T_B)$
\STATE \textbf{Initialize:} $samples \leftarrow \emptyset$

\FOR{each sample $s_i \in D$}
    \STATE Extract text $text_A \leftarrow s_i[T_A]$
    \STATE Extract text $text_B \leftarrow s_i[T_B]$
    \STATE Create sample: $(idx=2i, sentence=text_A, label=0)$
    \STATE Create sample: $(idx=2i+1, sentence=text_B, label=1)$
\ENDFOR

\STATE Shuffle sample indices randomly
\STATE \textbf{Output:} Dataset of size $2n$ with balanced 50-50 label distribution
\end{algorithmic}
\end{algorithm}
This section describes the comprehensive task data preparation methodology for evaluating paraphrase-based LLM detection systems. We present five distinct experimental settings that systematically vary data utilization strategies and task formulations to provide robust evaluation frameworks for different detection scenarios.

\paragraph{Rationale for Five-Setting Framework.} The five-setting evaluation framework addresses fundamental limitations in current AI text detection evaluation through systematic variation of three critical dimensions:\\
(1) \textbf{Data Utilization Strategy}: Exhaustive vs. sampling approaches to control semantic repetition;\\
(2) \textbf{Label Distribution}: Balanced vs. imbalanced scenarios to test base rate sensitivity;\\
(3) \textbf{Task Formulation}: Absolute vs. comparative classification paradigms. This comprehensive approach provides \textbf{convergent validity}—consistent performance across settings indicates robust detection capabilities, while performance divergence reveals specific vulnerabilities crucial for practical deployment.

\subsection{Setting 1: Single-Sentence Exhaustive Method.}
\label{appD:setup_1}
The exhaustive method implements a comprehensive data utilization strategy where all available instances from both relevant text types are used to create the maximum possible dataset size. Algorithm.\ref{algo:exhaustive_method} shows the technical specifics of creating such task data.

\paragraph{Characteristics.} Dataset size: $2 \times$ original size (e.g., 16,233 → 32,466 samples); Label distribution: Fixed 50-50 balance; Data utilization: Exhaustive use of all available instances; Semantic coverage: Maximum semantic diversity through complete enumeration.

\paragraph{Theoretical Motivation.} The exhaustive method embodies the principle of \textbf{maximum data utilization} for establishing performance upper bounds. This approach provides statistical power through larger datasets (32k samples), real-world representativeness (attackers can generate multiple paraphrases), and comprehensive coverage across all paraphrase variations. However, it faces the semantic similarity challenge: Type3 and Type4 both derive from Type1, potentially allowing models to exploit repeated semantic patterns rather than learning true stylistic discrimination, leading to evaluation inflation.

\subsection{Settings 2-4: Single-Sentence Sampling Method.} The sampling method addresses fundamental evaluation validity concerns by implementing \textbf{controlled semantic exposure}. This approach prevents models from exploiting repeated semantic patterns that could inflate performance metrics, ensuring evaluation focuses on true detection capabilities rather than content memorization. The method randomly samples only one instance per original sample, while allowing systematic control of label distribution through configurable sampling probabilities. Technical details can be represented by Algorithm.\ref{algo:sampling_method}.

\paragraph{Distribution Settings.} \textbf{Setting 2 (30-70):} Sampling probability $p = 0.3$ (30\% chance to sample Type B), expected distribution 30\% Label 1, 70\% Label 0, focusing on imbalanced dataset performance with minority LLM-generated content. \textbf{Setting 3 (50-50):} Sampling probability $p = 0.5$ (50\% chance for each type), balanced 50-50 distribution, enabling direct comparison with exhaustive method while eliminating semantic repetition. \textbf{Setting 4 (80-20):} Sampling probability $p = 0.8$ (80\% chance to sample Type B), expected distribution 80\% Label 1, 20\% Label 0, testing detector robustness under realistic scenarios with majority LLM-generated content.

\begin{algorithm}[H]
\caption{Single-Sentence Sampling Method}
\begin{algorithmic}[1]
\label{algo:sampling_method}
\STATE \textbf{Input:} Dataset $D$ with $n$ samples, Task $(T_A, T_B)$, sampling ratio $p$
\STATE \textbf{Initialize:} $samples \leftarrow \emptyset$, random seed

\FOR{each sample $s_i \in D$}
    \STATE Generate random value $r \sim \text{Uniform}(0,1)$
    \IF{$r < p$}
        \STATE Select $text \leftarrow s_i[T_B]$, $label \leftarrow 1$
    \ELSE
        \STATE Select $text \leftarrow s_i[T_A]$, $label \leftarrow 0$
    \ENDIF
    \STATE Create sample: $(idx=i, sentence=text, label=label)$
    \STATE $samples \leftarrow samples \cup \{(i, text, label)\}$
\ENDFOR

\STATE \textbf{Output:} Dataset of size $n$ with target label distribution
\end{algorithmic}
\end{algorithm}

\paragraph{Dynamic Label Distribution Rationale.} The three distribution settings (30\%, 50\%, 80\% machine-generated) address critical evaluation biases: \textbf{For Zero-Shot Detectors:} Base rate sensitivity (many metrics are sensitive to class imbalance), threshold robustness (optimal cutoff points may shift with prevalence), and calibration assessment (whether metric scores remain meaningful across varying base rates). \textbf{For Model-Based Detectors:} Prior assumption testing (implicit priors about AI text prevalence from training data), confidence calibration (reliability across different class distributions), and decision boundary stability (generalization across distribution shifts).

\subsection{Setting 5: Sentence-Pair Recognition.}
\label{appD:setup5}
Sentence pair recognition addresses a fundamental limitation in current AI text detection evaluation. Traditional single-sentence classification assumes detectors can establish absolute thresholds for "machine-likeness," but in practice, detection often involves \textbf{relative comparisons}. Pair-wise evaluation better mirrors real-world scenarios where humans and detectors must choose between alternatives of unknown provenance.

\paragraph{Evaluation Advantages}
\textbf{For Zero-Shot Detectors:} Eliminates threshold dependency (compare relative metric scores instead of learning optimal cutoffs), reduces calibration bias (pair-wise comparison within same semantic context normalizes domain/length variations), tests discriminative power directly (forces fine-grained distinctions between similar absolute scores).\\
\textbf{For Model-Based Detectors:} Mimics human judgment (natural comparative tasks vs. absolute classification), reduces prompt sensitivity (binary comparison prompts more stable than threshold-based), tests robustness (prevents exploitation of spurious correlations). This reveals whether detection capabilities stem from absolute text properties versus relative discriminative features—crucial for understanding detector reliability across domains and attack sophistication levels.

\begin{algorithm}[H]
\caption{Sentence-Pair Recognition Challenge}
\begin{algorithmic}[1]
\STATE \textbf{Input:} Dataset $D$ with $n$ samples, Task $(T_A, T_B)$
\STATE \textbf{Initialize:} $pairs \leftarrow \emptyset$

\FOR{each sample $s_i \in D$}
    \STATE Extract $sentence_A \leftarrow s_i[T_A]$
    \STATE Extract $sentence_B \leftarrow s_i[T_B]$
    \STATE Generate random bit $flip \sim \text{Bernoulli}(0.5)$
    
    \IF{$flip = 0$}
        \STATE $pair \leftarrow [sentence_A, sentence_B]$
        \STATE $labels \leftarrow [0, 1]$
    \ELSE
        \STATE $pair \leftarrow [sentence_B, sentence_A]$
        \STATE $labels \leftarrow [1, 0]$
    \ENDIF
    
    \STATE Create sample: $(idx=i, sentence\_pair=pair, label\_pair=labels)$
    \STATE $pairs \leftarrow pairs \cup \{(i, pair, labels)\}$
\ENDFOR

\STATE \textbf{Output:} Dataset of $n$ sentence pairs with randomized order
\end{algorithmic}
\end{algorithm}

\paragraph{Output Format and Applications.} Each sentence pair sample follows standardized format: \texttt{sentence\_pair} (tuple of two sentences for comparison), \texttt{label\_pair} (corresponding labels 0=human/original, 1=machine/modified), with order randomization preventing positional bias.\\ 
\textbf{Research Applications:} \\
\textbf{Zero-Shot Detection:} Compare metric values between sentence pairs, determine which scores higher on detection metrics, evaluate relative performance without absolute thresholds.\\
\textbf{Model-Based Approaches:} Prompt-tuned models for comparative judgments ("Which sentence is more likely to be machine-generated?"), comparative reasoning evaluation.\\ \textbf{Bias Analysis:} Study positional bias in sentence pair tasks, evaluate order-independence of detection systems, test robustness across different presentation formats.

\subsection{Summary}

To comprehensively evaluate paraphrase-based detection systems, we design a five-setting methodology that systematically varies three evaluation dimensions: data utilization strategy (exhaustive vs. sampling), label distribution (30-70, 50-50, 80-20), and task formulation (single-sentence vs. sentence-pair classification). This framework distinguishes robust detection capabilities from evaluation artifacts and assesses real-world deployment readiness.

\paragraph{Evaluation Settings.}
\begin{enumerate}[nosep]
    \item \textbf{Single-Sentence Exhaustive:} Uses all available samples with balanced 50-50 distribution
    \item \textbf{Single-Sentence Sampling (30-70):} Random sampling with 30\% positive, 70\% negative
    \item \textbf{Single-Sentence Sampling (50-50):} Random sampling with balanced distribution
    \item \textbf{Single-Sentence Sampling (80-20):} Random sampling with 80\% positive, 20\% negative
    \item \textbf{Sentence-Pair Recognition:} Pairwise comparison tasks with random order presentation
\end{enumerate}

\paragraph{Comparative Analysis.}
Table \ref{tab:setup_comparison} presents a systematic comparison of the five evaluation settings across seven critical dimensions. The exhaustive method provides maximum data utilization ($2n$ samples) but introduces semantic repetition concerns, while sampling methods eliminate repetition at the cost of reduced dataset size. Distribution variations (30-70, 50-50, 80-20) enable testing detector robustness across different base rates, with each setting targeting specific research focuses—minority class detection, balanced evaluation, or majority class scenarios. The sentence-pair setting uniquely provides comparative evaluation that eliminates threshold calibration dependencies but introduces potential positional bias concerns.

\begin{table*}[t]
\centering
\caption{Comparative analysis of the five evaluation settings.}
\label{tab:setup_comparison}
\begin{tabular}{@{}lcccccc@{}}
\toprule
\textbf{Aspect} & \textbf{Exhaustive} & \textbf{30-70} & \textbf{50-50} & \textbf{80-20} & \textbf{Sentence-Pair} \\
\midrule
Dataset Size & $2n$ & $n$ & $n$ & $n$ & $n$ \\
Label Distribution & 50-50 & 30-70 & 50-50 & 80-20 & Balanced pairs \\
Semantic Repetition & Present & Eliminated & Eliminated & Eliminated & Eliminated \\
Data Utilization & Complete & Sampled & Sampled & Sampled & Complete per pair \\
Evaluation Type & Absolute & Absolute & Absolute & Absolute & Comparative \\
Bias Concerns & Semantic & Distribution & Minimal & Distribution & Positional \\
Research Focus & Max performance & Minority class & Balanced & Majority class & Comparative \\
\bottomrule
\end{tabular}
\end{table*}

\paragraph{Key Advantages.}
This framework provides four critical evaluation capabilities: (1) \textbf{Robustness testing}—consistent performance across settings indicates robust detectors while divergence reveals vulnerabilities; (2) \textbf{Base rate sensitivity}—multiple distributions test reliability under varying real-world conditions; (3) \textbf{Content vs. style discrimination}—exhaustive vs. sampling comparison reveals whether detectors memorize content or detect stylistic patterns; (4) \textbf{Threshold independence}—sentence-pair evaluation eliminates calibration dependencies for zero-shot methods.

By testing performance consistency across these realistic deployment scenarios, the methodology reveals whether detection capabilities generalize beyond controlled laboratory conditions to real-world settings with varying base rates, semantic contexts, and attack sophistication levels.

\section{Detailed Evaluation Settings}

This section provides comprehensive details about the evaluation methodology, metrics, and experimental configurations used to assess the performance of various AI-generated text detection methods. Our evaluation framework encompasses both zero-shot detection methods and model-based approaches, tested across multiple challenging tasks designed to evaluate robustness against sophisticated text generation and paraphrasing attacks.

\subsection{Evaluation Metrics}

We employ a comprehensive set of evaluation metrics to assess detector performance across different aspects of AI-generated text detection. The metrics are designed to capture both binary classification performance and the ability to distinguish between human and machine-generated text under various conditions.

\textbf{Area Under ROC Curve (AUROC):} Measures the detector's ability to distinguish between human and AI-generated text across all classification thresholds. AUROC values range from 0.5 (random performance) to 1.0 (perfect classification).

\textbf{TPR@1\%FPR:} True Positive Rate when False Positive Rate is constrained to 1\%
\textbf{TPR@5\%FPR:} True Positive Rate when False Positive Rate is constrained to 5\%
\textbf{TPR@10\%FPR:} True Positive Rate when False Positive Rate is constrained to 10\%

These metrics are crucial for real-world deployment where false accusations of AI generation can have serious consequences.

\subsection{Zero-Shot Detectors Setup}
\label{appE:zero-shot_setup}
Zero-shot detectors require no training on the target detection task and rely on intrinsic properties of language models or statistical analysis of text characteristics. We evaluate four state-of-the-art zero-shot detection methods.

\subsubsection{Binoculars Configuration}

\textbf{Method Overview:} Binoculars leverages the observation that most decoder-only language models share substantial overlap in pretraining data, enabling cross-model probability comparison for detection.

\textbf{Model Configuration:}
\begin{enumerate}[nosep]
    \item \textbf{Observer Model:} \texttt{tiiuae/falcon-7b}
    \item \textbf{Performer Model:} \texttt{tiiuae/falcon-7b-instruct}
    \item \textbf{Detection Mode:} Accuracy-optimized (alternative: low-fpr mode)
\end{enumerate}

\textbf{Detection Process:}
\begin{enumerate}[nosep]
    \item Compute log probabilities using both observer and performer models
    \item Calculate Binoculars score based on probability discrepancy
    \item Apply global threshold (0.9015) for binary classification
    \item Alternative: Use model-specific thresholds for optimal performance
\end{enumerate}

\subsubsection{GTLR (Giant Language Model Test Room) Configuration}

\textbf{Method Overview:} GTLR analyzes token-level probability rankings and distributions to identify patterns characteristic of AI-generated text.

\textbf{Model Configuration:}
\begin{enumerate}[nosep]
    \item \textbf{Primary Model:} GPT-2-large
    \item \textbf{Analysis Window:} Full text sequences up to model maximum length
    \item \textbf{Probability Computation:} Token-wise conditional probabilities
\end{enumerate}

\textbf{Detection Features:}
\begin{enumerate}[nosep]
    \item \textbf{Rank Analysis:} Distribution of token ranks in vocabulary
    \item \textbf{Probability Patterns:} Statistical analysis of token probabilities
    \item \textbf{Entropy Measures:} Information-theoretic measures of text predictability
    \item \textbf{N-gram Statistics:} Higher-order linguistic pattern analysis
\end{enumerate}

\textbf{Threshold Selection:} Dynamic thresholding based on text length and domain characteristics, with fallback to empirically determined global thresholds.

\subsubsection{Fast-DetectGPT Configuration}

\textbf{Method Overview:} Fast-DetectGPT improves upon DetectGPT by using conditional probability curvature analysis, achieving 340× speedup while maintaining superior accuracy.

\textbf{Model Configuration:}
\begin{enumerate}[nosep]
    \item \textbf{Scoring Model:} \texttt{falcon-7b-instruct}
    \item \textbf{Sampling Model:} Same as scoring model for efficiency
    \item \textbf{Sampling Parameters:} 10,000 samples for probability estimation
\end{enumerate}

\textbf{Detection Algorithm:}
\begin{enumerate}[nosep]
    \item Compute log-likelihood of original text under scoring model
    \item Generate samples from reference model probability distribution
    \item Calculate mean and standard deviation of sample log-likelihoods
    \item Compute Fast-DetectGPT criterion:
    \begin{equation}
    \text{Criterion} = \frac{\log p_{\theta}(x) - \mu_{\tilde{x}}}{\sigma_{\tilde{x}}}
    \end{equation}
    where $\mu_{\tilde{x}}$ and $\sigma_{\tilde{x}}$ are sample statistics
    \item Apply threshold for binary classification (typically around 0.0)
\end{enumerate}

\subsubsection{RADAR Configuration}

\textbf{Method Overview:} RADAR (Robust AI-Text Detection via Adversarial Learning) uses adversarial training to achieve robustness against paraphrasing attacks.

\textbf{Model Configuration:}
\begin{enumerate}[nosep]
    \item \textbf{Base Model:} RoBERTa-large (355M parameters)
    \item \textbf{RADAR Model:} TrustSafeAI/RADAR-Vicuna-7B
    \item \textbf{Input Processing:} Maximum sequence length of 512 tokens
\end{enumerate}

\textbf{Detection Process:}
\begin{enumerate}[nosep]
    \item Tokenize input text using RoBERTa tokenizer
    \item Forward pass through adversarially trained model
    \item Apply log-softmax to output logits
    \item Extract probability of AI-generated class:
    \begin{equation}
    P(\text{AI-generated}) = \exp(\log\text{-softmax}(\text{logits})_0)
    \end{equation}
\end{enumerate}

\subsection{Model-Based Detectors Setup}

Model-based detectors require training on labeled datasets and can be categorized into traditional machine learning approaches and modern neural network-based methods.

\textbf{Method Overview:} Large language models are employed as detectors through carefully designed prompts and few-shot learning, leveraging their inherent understanding of text patterns.

\textbf{Model Selection:} Claude-3.5-Haiku, DeepSeek-V2.5, Gemma-3-27B, Llama-4-Scout-17B, Mistral-Nemo, Llama-4-Maverick-17B, WizardLM-2-8x22B

\textbf{Prompt Engineering Strategy:}
\begin{enumerate}
    \item \textbf{System Message:} Establishes expert persona and task context
    \item \textbf{Few-Shot Examples:} 3-4 carefully crafted demonstrations per task
    \item \textbf{Multi-Turn Conversation:} Maintains context across examples
    \item \textbf{Task-Specific Expertise:} Specialized personas for different detection tasks
\end{enumerate}

\paragraph{Single-sentence Classification Prompt Templates:}~\\
\label{appE:single-sentence_model_prompting}
\noindent \textbf{Task 1 - Paraphrase Source Attribution System Message:}
\begin{quote}
\itshape
``You are an expert text analyst specializing in paraphrase detection. Your task is to determine whether a paraphrased sentence was created by a human or by an AI/LLM system.

TASK CONTEXT: You are analyzing paraphrased versions of original text. Human paraphrases tend to be more natural, contextually aware, and show human linguistic intuition. LLM paraphrases often exhibit systematic patterns, over-formalization, or subtle unnaturalness.

CLASSIFICATION CRITERIA:
- Human paraphrase (0): Natural flow, contextual understanding, human-like word choices, appropriate informality/formality
- LLM paraphrase (1): Systematic rewording patterns, over-precision, unnatural phrasing, AI-like formalization

IMPORTANT: You must respond with ONLY the number 0 or 1. No explanation or additional text is allowed.''
\end{quote}

\textbf{Task 1 Few-Shot Examples:}
\begin{itemize}
    \item \textbf{Example 1 (LLM):} ``The expeditious mahogany-colored vulpine creature propels itself in a vertical trajectory above the lethargic canine'' → \textbf{Label: 1}
    \item \textbf{Example 2 (Human):} ``A quick brown fox jumps over a sleepy dog'' → \textbf{Label: 0}
    \item \textbf{Example 3 (LLM):} ``The research methodology employed in this investigation demonstrates a comprehensive approach to data collection and analysis'' → \textbf{Label: 1}
\end{itemize}

\noindent \textbf{Task 2 - General Text Authorship Detection System Message:}
\begin{quote}
\itshape
``You are an expert in AI-generated text detection. Your task is to determine whether a sentence was originally written by a human or generated by an AI/LLM system.

TASK CONTEXT: You are analyzing original text authorship. Human-written text shows natural creativity, personal voice, and authentic expression. LLM-generated text often exhibits training patterns, generic phrasing, or artificial coherence.

CLASSIFICATION CRITERIA:
- Human original (0): Authentic voice, natural imperfections, personal style, genuine creativity, contextual authenticity
- LLM generated (1): Training data patterns, generic expressions, artificial smoothness, systematic structure

IMPORTANT: You must respond with ONLY the number 0 or 1. No explanation or additional text is allowed.''
\end{quote}

\textbf{Task 2 Few-Shot Examples:}
\begin{itemize}
    \item \textbf{Example 1 (LLM):} ``In today's rapidly evolving digital landscape, organizations must leverage cutting-edge technologies to optimize their operational efficiency and drive sustainable growth.'' → \textbf{Label: 1}
    \item \textbf{Example 2 (Human):} ``Ugh, my coffee maker broke again this morning. Third time this month! I swear these things are designed to fail right after the warranty expires.'' → \textbf{Label: 0}
    \item \textbf{Example 3 (LLM):} ``The implementation of artificial intelligence solutions presents numerous opportunities for enhancing productivity while simultaneously addressing potential challenges related to workforce adaptation.'' → \textbf{Label: 1}
\end{itemize}

\noindent \textbf{Task 3 - AI Text Laundering Detection System Message:}
\begin{quote}
\itshape
``You are a specialist in detecting AI text laundering techniques. Your task is to determine the level of AI processing in text — distinguishing between LLM-paraphrased original content versus LLM-paraphrased generated content.

TASK CONTEXT: You are comparing two types of AI-processed text: (1) LLM paraphrases of human original text, and (2) LLM paraphrases of AI-generated text. The second type represents deeper AI processing and "laundering" attempts.

CLASSIFICATION CRITERIA:
- LLM paraphrased original (0): AI paraphrase of human content — retains some human authenticity beneath AI processing
- LLM paraphrased generated (1): AI paraphrase of AI content — multiple layers of AI processing, more artificial patterns

IMPORTANT: You must respond with ONLY the number 0 or 1. No explanation or additional text is allowed.''
\end{quote}

\noindent \textbf{Task 4 - Iterative Paraphrase Depth Detection System Message:}
\begin{quote}
\itshape
``You are an expert in detecting iterative AI processing depth. Your task is to determine whether text has undergone fewer or more iterations of LLM paraphrasing.

TASK CONTEXT: You are analyzing text that has been paraphrased multiple times by AI systems. Earlier iterations retain more original characteristics, while deeper iterations show increasing AI processing artifacts and departure from natural expression.

CLASSIFICATION CRITERIA:
- 1st iteration paraphrase (0): Less deep AI processing — some original patterns remain, moderate AI influence
- 3rd iteration paraphrase (1): Deeper AI processing — heavily processed, multiple layers of AI transformation, more artificial

IMPORTANT: You must respond with ONLY the number 0 or 1. No explanation or additional text is allowed.''
\end{quote}

\noindent \textbf{Task 5 - Deep Paraphrase Attack Detection System Message:}
\begin{quote}
\itshape
``You are a cybersecurity expert specializing in detecting sophisticated AI paraphrase attacks. Your task is to distinguish between authentic human-written text and heavily processed AI paraphrases designed to evade detection.

TASK CONTEXT: You are facing the most challenging detection scenario — authentic human original text versus 3rd-iteration LLM paraphrases (the most sophisticated paraphrase attacks). These attacks are designed to fool detection systems.

CLASSIFICATION CRITERIA:
- Human original (0): Authentic human expression, natural imperfections, genuine voice, unprocessed authenticity
- Deep paraphrase attack (1): Heavily processed AI text, multiple transformation layers, sophisticated evasion attempts

IMPORTANT: You must respond with ONLY the number 0 or 1. No explanation or additional text is allowed.''
\end{quote}

\paragraph{Sentence-Pair Comparative Prompting:}~\
\label{appE:sentence_pair_model_prompting}
\noindent For sentence-pair tasks, the prompting strategy shifts to comparative analysis where models must determine which sentence in a pair exhibits more human-like or AI-like characteristics. The system messages are adapted to emphasize comparative judgment:

\noindent\textbf{Task 1 - Paraphrase Source Attribution (Sentence-Pair) System Message:}
\begin{quote}
\itshape``You are an expert text analyst specializing in comparative paraphrase detection. Your task is to determine which sentence in a pair was paraphrased by a human versus an AI/LLM system.

TASK CONTEXT: You are comparing two paraphrased versions of the same content. One was created by a human, the other by an AI/LLM. Human paraphrases show natural linguistic intuition, while LLM paraphrases exhibit systematic patterns.

COMPARISON CRITERIA:
- Human paraphrase: Natural flow, contextual awareness, human-like word choices, appropriate style
- LLM paraphrase: Systematic rewording, over-precision, unnatural phrasing, AI-like formalization

INSTRUCTIONS: 
1. Analyze both sentences carefully
2. Determine which sentence shows more human-like paraphrasing characteristics
3. Respond with 0 if the FIRST sentence is more human-like, 1 if the SECOND sentence is more human-like

IMPORTANT: You must respond with ONLY the number 0 or 1. No explanation or additional text is allowed.''
\end{quote}

\noindent\textbf{Task 2 - General Text Authorship Detection (Sentence-Pair) System Message:}
\begin{quote}
\itshape``You are an expert in comparative AI-generated text detection. Your task is to determine which sentence in a pair was originally written by a human versus generated by an AI/LLM system.

TASK CONTEXT: You are comparing original text authorship between two sentences. One is authentic human writing, the other is AI-generated. Human writing shows personal voice and natural expression, while AI writing exhibits training patterns.

COMPARISON CRITERIA:
- Human original: Authentic voice, natural imperfections, personal style, genuine creativity
- LLM generated: Training patterns, generic expressions, artificial smoothness, systematic structure

INSTRUCTIONS:
1. Analyze both sentences for authenticity markers
2. Determine which sentence shows more human authorship characteristics
3. Respond with 0 if the FIRST sentence is more human-authored, 1 if the SECOND sentence is more human-authored

IMPORTANT: You must respond with ONLY the number 0 or 1. No explanation or additional text is allowed.''
\end{quote}

\noindent\textbf{Task 3 - AI Text Laundering Detection (Sentence-Pair) System Message:}
\begin{quote}
\itshape ``You are a specialist in comparative AI text laundering detection. Your task is to determine which sentence in a pair shows deeper AI processing - comparing LLM-paraphrased original content versus LLM-paraphrased generated content.

TASK CONTEXT: You are comparing two types of AI-processed text to identify which has undergone more intensive AI processing (text laundering). One is an LLM paraphrase of human content, the other is an LLM paraphrase of AI content.

COMPARISON CRITERIA:
- LLM paraphrased original: AI processing of human content - retains some authenticity
- LLM paraphrased generated: AI processing of AI content - deeper artificial patterns, more laundering

INSTRUCTIONS:
1. Analyze both sentences for depth of AI processing
2. Determine which sentence shows more intensive AI laundering
3. Respond with 0 if the FIRST sentence shows more AI laundering, 1 if the SECOND sentence shows more AI laundering

IMPORTANT: You must respond with ONLY the number 0 or 1. No explanation or additional text is allowed.''
\end{quote}

\noindent\textbf{Task 4 - Iterative Paraphrase Depth Detection (Sentence-Pair) System Message:}
\begin{quote}
\itshape``You are an expert in comparative iterative AI processing analysis. Your task is to determine which sentence in a pair has undergone deeper iterative LLM paraphrasing.

TASK CONTEXT: You are comparing sentences that have been paraphrased different numbers of times by AI systems. One has fewer iterations, the other has more iterations. Deeper iterations show increasing AI processing artifacts.

COMPARISON CRITERIA:
- 1st iteration: Less deep processing - some original characteristics remain
- 3rd iteration: Deeper processing - heavily transformed, more artificial patterns

INSTRUCTIONS:
1. Analyze both sentences for depth of iterative processing
2. Determine which sentence shows more iterations of AI paraphrasing
3. Respond with 0 if the FIRST sentence shows deeper processing, 1 if the SECOND sentence shows deeper processing

IMPORTANT: You must respond with ONLY the number 0 or 1. No explanation or additional text is allowed.''
\end{quote}

\noindent\textbf{Task 5 - Deep Paraphrase Attack Detection (Sentence-Pair) System Message:}
\begin{quote}
\itshape``You are a cybersecurity expert specializing in comparative detection of sophisticated AI paraphrase attacks. Your task is to distinguish between authentic human-written text and heavily processed AI paraphrases in direct comparison.

TASK CONTEXT: You are facing the ultimate detection challenge - comparing authentic human original text against 3rd-iteration LLM paraphrases (sophisticated evasion attacks) to identify which is the authentic human text.

COMPARISON CRITERIA:
- Human original: Authentic expression, natural imperfections, genuine voice, unprocessed
- Deep paraphrase attack: Heavily processed, multiple AI transformations, sophisticated evasion

INSTRUCTIONS:
1. Analyze both sentences for authenticity versus AI processing
2. Determine which sentence is the authentic human original
3. Respond with 0 if the FIRST sentence is more human-original, 1 if the SECOND sentence is more human-original

IMPORTANT: You must respond with ONLY the number 0 or 1. No explanation or additional text is allowed.''
\end{quote}

\noindent \textbf{Multi-Turn Conversation Structure:}

Each evaluation follows a consistent multi-turn conversation pattern:
\begin{enumerate}
    \item \textbf{System Message:} Task-specific expert persona and classification criteria
    \item \textbf{Few-Shot Example 1:} User query with example text → Assistant response with label
    \item \textbf{Few-Shot Example 2:} User query with example text → Assistant response with label  
    \item \textbf{Few-Shot Example 3:} User query with example text → Assistant response with label
    \item \textbf{Target Query:} User query with actual text to classify → Assistant response (evaluated)
\end{enumerate}

This structure ensures consistent context establishment and provides clear behavioral examples before the actual classification task.

\end{document}